%% file: acl2023.tex
\pdfoutput=1

\documentclass[11pt]{article}
\usepackage{overpic}
\usepackage{float}

\newif\ifcomment\commenttrue
\input{style/preamble}

\usepackage[]{acl2023}

\usepackage{xspace}
\usepackage{xcolor}
\usepackage[utf8]{inputenc}
\usepackage{pgfplots}
\usepackage{dsfont}
\DeclareUnicodeCharacter{2212}{−}
\usepgfplotslibrary{groupplots,dateplot}
\usetikzlibrary{patterns,shapes.arrows}
\pgfplotsset{compat=newest}

\usepackage{tikzscale}
\usepackage{relsize}
\usepackage{amsmath,amssymb}
\newcommand{\probP}{\text{I\kern-0.15em P}}

\usepackage{multirow, colortbl}

\usepackage{tabularx,booktabs}
\usepackage{makecell}

\usepackage[normalem]{ulem}
\useunder{\uline}{\ul}{}

\usepackage{graphicx}

\definecolor{ablation6}{HTML}{fcefed}
\definecolor{ablation_tie}{HTML}{fce3e1}

\definecolor{ablation5}{HTML}{fcd8d4}
\definecolor{ablation4}{HTML}{FBC3BC}
\definecolor{ablation3}{HTML}{F7A399}
\definecolor{ablation2}{HTML}{F38375}
\definecolor{ablation1}{HTML}{EF6351}

\definecolor{OliveGreen}{rgb}{0.05, 0.75, 0.24}
\definecolor{BrickRed}{rgb}{0.8, 0.25, 0.33}

\usepackage[]{algpseudocode}
\usepackage[]{algorithm}
\usepackage{float}
\algtext*{EndFor}%
\algtext*{EndProcedure}%

\usepackage{booktabs}
\usepackage[normalem]{ulem}
\useunder{\uline}{\ul}{}

\usepackage{times}
\usepackage{latexsym}
\usepackage{adjustbox}

\usepackage[T1]{fontenc}
\usepackage{amsmath}
\usepackage{amssymb}
\usepackage{booktabs}
\usepackage{tikzscale}
\usepackage{amsmath}

\usepackage{multirow, colortbl}

\usepackage{tabularx,booktabs}
\usepackage{makecell}
\usepackage{multirow}
\usepackage{scalerel,xparse}

\usepackage{cleveref}
\usepackage{microtype}
\usepackage[most]{tcolorbox}

\usepackage{enumitem}
\crefformat{section}{\S#2#1#3}
\crefformat{subsection}{\S#2#1#3}
\crefformat{subsubsection}{\S#2#1#3}

\definecolor{bggray}{rgb}{0.95, 0.95, 0.95}
\usepackage[%
    framemethod=tikz,
    skipbelow=\topskip,
    skipabove=\topskip
]{mdframed}
\mdfsetup{%
    leftmargin=0pt,
    rightmargin=0pt,
    backgroundcolor=bggray,
    middlelinecolor=black,
    roundcorner=3
}

\definecolor{SkyBlue}{rgb}{0.53, 0.81, 0.92}

\newtcolorbox[
  list inside=prompt,
  auto counter,
  number within=section
]{prompt}[1][]{%
  enhanced,
  float*=t, 
  colbacktitle=black!60,
  fonttitle=\small,
  coltitle=white,
  fontupper=\footnotesize,
  boxsep=4pt,
  left=0pt, right=0pt, top=0pt, bottom=0pt,
  boxrule=1pt,
  width=\textwidth,          
  enlarge left by=0mm,
  enlarge right by=0mm,
  listing only,
  listing options={
    basicstyle=\ttfamily\footnotesize,
    breaklines=true,
    breakatwhitespace=true,
    language=json
  },
  #1,
}

\newtcolorbox[
  list inside=trace,
  auto counter,
  number within=section
]{trace}[1][]{%
  enhanced,
  float*=t,
  colback=blue!5,             
  colbacktitle=blue!60!black, 
  colframe=blue!60!black,     
  fonttitle=\small,
  coltitle=white,
  fontupper=\footnotesize,
  boxsep=4pt,
  left=0pt, right=0pt, top=0pt, bottom=0pt,
  boxrule=1pt,
  width=\textwidth,
  enlarge left by=0mm,
  enlarge right by=0mm,
  listing only,
  listing options={
    basicstyle=\ttfamily\footnotesize,
    breaklines=true,
    breakatwhitespace=true,
    language=json
  },
  #1,
}

\definecolor{UMDred}{HTML}{ed1c24}

\definecolor{yellowcolor}{HTML}{ffc20e}
\definecolor{redcolor}{HTML}{e99999}
\definecolor{orangecolor}{HTML}{f6b26b}
\definecolor{yellowcolor}{HTML}{ffd966}
\definecolor{bluecolor}{HTML}{a0c5e8}
\definecolor{purplecolor}{HTML}{d9d2e9}

\usepackage{xcolor}
\definecolor{highlight}{HTML}{FFAE02}

\title{Check The Scoreboard:\\
An Analysis of Scoring Schemes on Multiple-Choice Evaluation}

\newcommand{\authorSpacing}{0.5cm}
\author{
\textbf{Nishant Balepur}$^{1, 2}$\hspace{\authorSpacing}
\textbf{Paiheng Xu}$^{1}$\\
\textbf{Wei Ai}$^{1}$ \hspace{\authorSpacing}
\textbf{Eunsol Choi}$^{2}$ \hspace{\authorSpacing}
\textbf{Rachel Rudinger}$^{1}$ \hspace{\authorSpacing}
\textbf{Jordan Boyd-Graber}$^{3}$\footnotemark \\[0.5em]
$^{1}$University of Maryland \hspace{0.3cm}
$^{2}$New York University \hspace{0.3cm}
$^{3}$Nanyang Technological University \hspace{0.3cm}
\\[0.5em]
\texttt{nbalepur@umd.edu} \hspace{0.5em} \texttt{jordan.ying@ntu.edu.sg}
}

\begin{document}
\maketitle

\input{2026_arr_mcqa_scoring/sections/00_abstract}

\input{2026_arr_mcqa_scoring/sections/10_intro}

\input{2026_arr_mcqa_scoring/sections/20_method}

\input{2026_arr_mcqa_scoring/sections/30_results}

\input{2026_arr_mcqa_scoring/sections/50_related_work}

\input{2026_arr_mcqa_scoring/sections/60_conclusion}

\input{2026_arr_mcqa_scoring/sections/70_limitation_ethics}

\bibliography{custom}
\bibliographystyle{acl_natbib}

\clearpage

\appendix
\input{2026_arr_mcqa_scoring/sections/100_appendix}

\end{document}

%% file: style/preamble.tex
\usepackage[a-1b]{pdfx}

\usepackage{framed}
\usepackage{lmodern}
\usepackage{mdwlist}
\usepackage{siunitx}
\usepackage{latexsym}
\usepackage{colortbl}
\usepackage{xcolor}
\usepackage{nicefrac}
\usepackage{booktabs}
\usepackage{fnpct}
\usepackage{amsfonts}
\usepackage[T1]{fontenc}
\usepackage{bold-extra}
\usepackage{amsmath}
\usepackage{amssymb}
\usepackage{bm}
\usepackage{graphicx}
\usepackage{mathtools}
\usepackage{microtype}
\usepackage{multirow}
\usepackage{multicol}
\usepackage{xpatch}
\usepackage{latexsym,comment}
\usepackage[normalem]{ulem}

\newcommand*{\missingreference}{{\Huge \colorbox{red}{?reference?}}}
\newcommand*{\missingcitation}{{\Huge \colorbox{red}{?citation?}}}

\newcommand{\Norm}{\abr{NR}}
\newcommand{\NormVar}{\abr{NR-Var}}
\newcommand{\NM}{\abr{NM}}
\newcommand{\Elim}{\abr{Elim}}
\newcommand{\ElimNM}{\abr{Elim-NM}}
\newcommand{\CBM}{\abr{CBM}}
\newcommand{\PPA}{\abr{PPA}}
\newcommand{\AUC}{\abr{AUC}}

\makeatletter
\xpatchcmd{\@setref}{\bfseries}{\missingreference}{}{}
\def\@citex[#1]#2{\leavevmode
    \let\@citea\@empty
    \@cite{\@for\@citeb:=#2\do
        {\@citea\def\@citea{,\penalty\@m\ }%
            \edef\@citeb{\expandafter\@firstofone\@citeb\@empty}%
            \if@filesw\immediate\write\@auxout{\string\citation{\@citeb}}\fi
            \@ifundefined{b@\@citeb}{\hbox{\reset@font\missingcitation}%
                \G@refundefinedtrue
                \@latex@warning
                {Citation `\@citeb' on page \thepage \space undefined}}%
            {\@cite@ofmt{\csname b@\@citeb\endcsname}}}}{#1}}
\makeatother

\newcommand{\mm}[0]{\textsc{llm}\xspace}
\newcommand{\mcqa}{\textsc{mcqa}\xspace}

\newcommand{\mcq}{\textsc{mcq}\xspace}

\newcommand{\gem}[1]{\mbox{\textsc{gem}}}
\newcommand{\abr}[1]{\textsc{#1}\xspace}

\newcommand{\hidetext}[1]{}
\newcommand{\ignore}[1]{}

\ifcomment
    \newcommand{\pinaforecomment}[3]{\colorbox{#1}{\parbox{.8\linewidth}{#2: #3}}}

    \newcommand{\prtodo}[1]{\pinaforecomment{lightblue}{pr}{#1}}
    \newcommand{\prtodoi}[1]{\pinaforecomment{lightblue}{pr}{#1}}
\else
    \newcommand{\pinaforecomment}[3]{}
    \newcommand{\prtodo}[1]{}
    \newcommand{\prtodoi}[1]{}
\fi

\newcommand{\smallurl}[1]{ \begin{tiny}\url{#1}\end{tiny}}

\definecolor{lightblue}{HTML}{3cc7ea}
\definecolor{CUgold}{HTML}{CFB87C}
\definecolor{grey}{rgb}{0.95,0.95,0.95}
\definecolor{ceil}{rgb}{0.57, 0.63, 0.81}
\definecolor{UMDred}{HTML}{ed1c24}
\definecolor{UMDyellow}{HTML}{ffc20e}

\newcommand{\nlp}[0]{\abr{nlp}}


%% file: 2026_arr_mcqa_scoring/sections/00_abstract.tex
\begin{abstract} {

Multiple-choice question answering (\mcqa{}) benchmarks in \nlp{} use number-right scoring (accuracy), but in educational testing, the scoring scheme---the combination of the response mode models follow and the rule for grading responses---is a key design choice that dictates which abilities to reward.
We examine how alternatives to number right change what \mcqa{} measures with six education-inspired schemes that assess abilities beyond accuracy: distractor elimination, abstention, confidence calibration, and self-correction.
On \mm{} benchmarks, these schemes:
1) shift rankings of 31 \mm{}s beyond rephrased number right prompts;
2) better predict the \mm{}s users prefer in LLM Arena; and
3) reveal distinct model capabilities, like that GPT-5 rarely abstains and readily self-corrects, while weaker open-weight models often abstain and hesitate to eliminate choices.~Given the benefits of alternative scoring schemes, we discuss ways to extend them to tasks beyond~\mcqa{}.\footnote{Our code: https://github.com/nbalepur/mcqa-scoring/}
\footnotetext{Work done at University of Maryland}
}
\end{abstract}

%% file: 2026_arr_mcqa_scoring/sections/10_intro.tex
\input{data/scoring}

\section{Introduction: Keeping Score in \mcqa{}} \label{section:intro}

In multiple-choice question answering \cite[\mcqa{}]{clark2020f} benchmarks, \nlp{} researchers carefully study question and inference design \cite{gu2024olmes},~but stick to the simplest \textbf{scoring scheme} of \textit{number right scoring}: models receive one point for each correct answer (i.e., accuracy).
Despite longstanding concerns that this rewards guessing~\citep{10.1145/3596490} and poorly reflects abilities~users value \citep{saxon2024benchmarks}, it persists due to its simplicity.

Conversely, education research offers alternative schemes that punish guessing, elicit confidence, or add re-attempts \cite{lau2011guessing}.
These~schemes surface student abilities that number right scoring cannot distinguish \cite{kanzow2023scoring}, but add costs like test anxiety that limit adoption \cite{ndu2016negative}.
However, these schemes~test abilities \mm{}s lack---confidence, abstention, and self-correction---and \mm{}s do not experience test anxiety \cite{10.1145/3624724}, making \nlp{} well-positioned to adapt them.

Our paper presents six \mcqa{} scoring schemes for \mm{} evaluation that use alternative prompt instructions and metrics (Table~\ref{table:scoring}).
Grounded~in~education research, our schemes assess how well~\mm{}s \textit{abstain} when uncertain to avoid penalties, use \textit{partial knowledge} to eliminate distractors, express calibrated \textit{confidence} in predictions, and \textit{self-correct} after being informed they~are wrong---distinct abilities that \mcqa{} neglects with number right scoring.

Across 31 \mm{}s and three \mcqa{} datasets~(\cref{section:experiments}), many schemes shift \mm{} ranks more than~rephrasing instructions in our number right prompt, testing unique abilities~(\cref{subsection:ranks}).
Answer until correct scoring best predicts~the~\mm{}s that users prefer~in LLM Arena \citep{zheng2024judging},~helping \mcqa{}~better reflect what users value (\cref{subsection:preferences}).
Our schemes~also reveal \mm{} behaviors: 
larger \mm{}s with test-time reasoning are more consistent over schemes
(\cref{subsection:consistency}), 
strong GPT-5 models rarely abstain and are quick to change answers, 
Command-R+ often~abstains, and LLaMA-3.1 hesitates to eliminate incorrect~choices (\cref{subsection:behavior})---quirks that number right fails to surface. 

These scoring schemes require only prompt and metric changes, making them simple to incorporate into existing \mcqa{} benchmarks.
We more~broadly argue for the importance of scoring scheme design in benchmarks and conclude with~ways to~extend them beyond \mcqa{} (\cref{section:conclusion}).
Our contributions~are:\\
\noindent \textbf{1)} Six education-grounded scoring schemes, evaluated across three \mcqa{} benchmarks and 31~\mm{}s.\\
\noindent \textbf{2)} Evidence that scoring schemes test diverse abilities and better predict model ranks in LLM Arena. \\
\noindent \textbf{3)} Cross-scheme analyses of consistency, measurements, confidence, and self-correction behaviors.

%% file: data/scoring.tex
\newcommand{\redhl}[1]{\colorbox{red!25}{#1}}
\newcolumntype{M}[1]{>{\raggedright\arraybackslash}m{#1}}

\begin{table*}[!h]
\scriptsize
\centering
\setlength{\tabcolsep}{3.8pt}
\begin{tabular}{@{}M{3.5cm}M{2.2cm}M{9.8cm}@{}}
\toprule
\textbf{Name} & \textbf{Abilities Tested} & \textbf{Description} \\
\midrule
Number Right \cite[\Norm]{kelly1916kansas} & Answer Accuracy & The standard method where the model picks a best answer. +1 point if correct and 0 if incorrect. \\
\midrule
Control Variation (\NormVar) & Prompt Sensitivity & To set a baseline prompt sensitivity when testing different schemes, we rephrase the prompt of \Norm. \\
\midrule
Negative Marking \cite[\NM]{holt2006analysis} & Uncertainty, Abstention & The model picks one best answer or abstains. To penalize guessing, scoring awards +1 point for right answers, $-1/(n-1)$ for wrong ones, and 0 for abstaining, so the expected value of guessing is 0. \\
\midrule
Elimination Scoring~\cite[\Elim]{coombs1956assessment} & Partial Knowledge, Eliminative Reasoning & The model picks as many answers it thinks are wrong. For granular partial knowledge, the score is the proportion of all distractors eliminated, or $0$ points if the correct answer is eliminated. \\
\midrule
Elimination + NM \cite[\ElimNM]{coombs1956assessment} & Partial Know., Elim. Reasoning, Abstention & A scoring scheme that combines the benefits and abilities tested in \Elim and \NM: models receive $1 / (n-1)$ points for eliminating wrong answers and $-1$ for correct ones. \\
\midrule
Confidence-Based Marking \cite[\CBM]{gardner2006confidence} & Discrete Confidence Calibration & The model predicts one answer choice and a confidence level ($C=1$ for low, $C=2$ for medium, or $C=3$ for high). To reward calibration, predicting $C=1$ yields $+1$ if correct and $0$ if wrong, $C=2$ yields $+2$ if correct and $-2$ if wrong, and $C=3$ yields $+3$ if correct and $-6$ if wrong. \\
\midrule
Personal Point Allocation \cite[\PPA]{macneil2023examining} & Probabilistic Conf. Calibration & The model outputs a probability distribution for each choice being correct. To reward confidently right answers, we use Brier score between the model's distribution and the true distribution (one-hot). \\
\midrule
Answer Until Correct \cite[\AUC]{wilcox1982some} & Self-Correction, Adaptability & Over a fixed number of tries, the model answers until correct---told when incorrect. To reward fewer attempts, the score is $(r-1)/(n-1)$, where $r$ is \# choices left when the correct answer is picked. \\
\bottomrule
\end{tabular}
\caption{\label{table:scoring} The \mcqa{} scoring schemes from education we adopt and abilities they test. Appendix~\ref{appendix:prompt} has examples.}
\end{table*}

%% file: 2026_arr_mcqa_scoring/sections/20_method.tex
\section{Experimental Setup: Scoring Schemes} \label{section:experiments}

Drawing on \citet{koele1987scoring}, a \underline{\textbf{scoring scheme}} combines a \textit{response mode} (how test-takers~must respond to inputs) and a \textit{scoring rule} (how~to~score responses).
Practically in \mm{} evaluation, prompts define response modes (e.g., ``\textit{Select the correct answer}'') and metrics (e.g., accuracy) dictate scoring.

\mcqa{}'s default scoring scheme is number~right \cite[\Norm]{kelly1916kansas}: given a question~$q$ and~$n$~choices $\mathcal{C}$, test-takers predict one answer $\hat{a} \in \mathcal{C}$---the \textit{response mode}.
The \textit{scoring rule} assigns one~point if $\hat{a}$ matches the gold answer ($\mathds{1}(a = \hat{a})$).
Despite its popularity, \Norm{} cannot distinguish lucky guesses from true understanding and provides no credit for test-takers' partial knowledge \cite{lau2011guessing}, which also applies to \nlp{} models \citep{10.1145/3596490}.

To see whether alternative scoring schemes~can bolster \mm{} evaluation, we study \mcqa{}~schemes inspired by education (Table~\ref{table:scoring}): negative marking (\NM) to penalize guessing \cite{Frary1988FormulaSO}, answer elimination (\Elim, \ElimNM) for partial credit based on the number of ruled-out choices \cite{coombs1956assessment}, confidence elicitation (\CBM, \PPA) to reward calibrated certainty \cite{finetti1965methods}, and answering until correct (\AUC) to evaluate how test-takers adapt to feedback \cite{wilcox1982some}.
These test desired abilities of \mm{}s that \Norm{} cannot capture \cite{liang2023holistic}---abstention \cite{elhady-etal-2025-wicked}, eliminative reasoning \cite{ma2023poe}, calibration \cite{guo2017calibration}, and self-correction \cite{huang2023large}---valuable additions to \mcqa{}.

We implement each scheme via custom prompts outlining response modes, and metrics as scoring rules.
To illustrate, for \NM, we adapt \Norm's prompt to let \mm{}s abstain as part of the response mode, and the scoring rule to add penalties for wrong answers (Table~\ref{table:scoring}, row 3).
We repeat~this process for each row in Table \ref{table:scoring}---prompting \mm{}s to return wrong choices in a list for \Elim/\ElimNM, predictions with confidence scores in \CBM/\PPA, and add past predictions as inputs for \AUC---specified in Appendix~\ref{appendix:prompt}.
All prompts use \mcqa{} templates drawn from the InspectAI library \cite{inspect_ai_framework}.


\subsection{Datasets} \label{subsection:dataset}

We sample $1000$ \mcq{}s from three popular benchmarks of varied difficulty: \textbf{1) ARC}--grade-school scientific knowledge/commonsense \cite{clark2018think};
\noindent \textbf{2) MMLU}---knowledge of 57 college topics \cite{hendrycks2020measuring}; and \textbf{3) SuperGPQA}---knowledge of 285 graduate topics \cite{du2025supergpqa}.

\subsection{Models} \label{subsection:model}

We~assess 31 frontier \mm{}s:
1) \textit{Gemini} \cite[2/3 Flash, 2.5 Lite, 2.5/3 Pro]{comanici2025gemini}; 2) \textit{GPT} \cite[4.1, 5 Nano, 5 Mini, 5, 5.2]{singh2025openai}; 3) \textit{Claude} \cite[3.7/4/4.5 Sonnet, 4.5 Haiku]{anthropic2025claude_sonnet_4_5}; 4) \textit{Cohere} Command-R \cite[R, R+, R7B]{cohere2024command_r}; 5) \textit{DeepSeek} \cite[V3.1, R1]{guo2025deepseek}; 6) \textit{Qwen}-3 \cite[80B Inst/Think, 235B Inst/Think]{yang2025qwen3}; 7) \textit{LLaMA}-3.1 \cite[8B, 70B, 405B]{dubey2024llama}; 8) \textit{GPT OSS} \cite[20B, 120B]{agarwal2025gpt}; 9) \textit{GLM} \cite[4.5, 4.6]{zeng2025glm}; and 10) \textit{Kimi}-K2 \cite{team2025kimi}.
GPT-OSS and Cmd-R+ take a week on SuperGPQA, so we omit them.
We use default model parameters.

%% file: 2026_arr_mcqa_scoring/sections/30_results.tex
\section{Results: Analyzing Scoring Schemes} \label{section:results}

Having designed education-based \mcqa{} schemes (\cref{section:experiments}), we now reveal their benefits for \nlp{}: they reward new abilities (\cref{subsection:ranks}), predict user preferences (\cref{subsection:preferences}), and surface model behaviors (\cref{subsection:consistency}, \cref{subsection:behavior}).

\subsection{Different Schemes, Different Ranks} \label{subsection:ranks}

While our alternative scoring schemes differ from \Norm by design, we do not need them if they measure the same model abilities as \Norm, as their adoption would never impact downstream evaluation \cite{cronbach1955construct}.
To probe this, we measure the Spearman's correlation between \mm{} rankings under each scheme versus \Norm{}.
Low correlation indicates discriminant validity \cite{campbell1959convergent}: the scheme evaluates abilities \Norm{} does not.


\NM and \CBM have high rank correlation with \Norm (Fig~\ref{fig:rankings}), but \Elim, \ElimNM, \PPA, and~\AUC consistently show lower correlation than~the \NormVar control, testing skills \Norm does not capture~beyond rephrased instructions.
This trend also holds on ARC when shuffling choices and replacing letters with numbers (e.g., (A) $\rightarrow$ (1)) as sensitivity baselines in Appendix Table~\ref{table:robustness}.
There is not a~best scheme, as this depends on what researchers value: \Elim for eliminative reasoning, \PPA for calibration, and \AUC~for self-correction (Table~\ref{table:scoring}).
\Norm cannot target such skills, which~have downstream use: medical chatbots apply \Elim in diagnosis by exclusion \cite{fred2013diagnosis}, while \abr{ai} tutors~apply \AUC when adapting to student errors \cite{chen2024multi}.

\input{figures/correlation}

For researchers who prefer the traditional \Norm scheme, applying many schemes can still break ties: ARC is a simpler dataset where \mm{}s score closely with \Norm{} (Appendix~\ref{appendix:ranks}), but \ElimNM~and \Elim have lower correlation with \Norm.
Thus, they can better separate \mm{} abilities as tie-breaking decisions.

\input{figures/llm_arena}
\input{figures/consistency}

\input{figures/behaviors}

\subsection{AUC Agrees with Human Preferences} \label{subsection:preferences}




While user feedback is the north-star goal for \mm{} development \citep{NEURIPS2022_b1efde53}, researchers often rely on cheaper offline proxies such as \mcqa{} to guide training \cite{olmo2025olmo}. 
To test~whether alternative scoring schemes improve \mcqa{} as an online proxy, we compare \mm{} ranks under each scheme to LLM Arena ranks---a large-scale benchmark of user preferences via live voting that~developers actively optimize for \cite{spangher-etal-2025-chatbot}.

\AUC~best agrees with LLM Arena (Figure~\ref{fig:llm_arena})---more than \Norm---so \AUC can better predict \mm{}s users prefer pre-deployment.
This matches \citet{shi-etal-2024-life} and \citet{liu-etal-2025-user} who show users prefer \mm{}s that adapt well to feedback, which~\AUC{} captures by making \mm{}s self-correct; this trend holds in a scheme where \mm{}s always self-reflect (Appendix~\ref{appendix:llm_arena}). 
As expected, the gap between~\Norm's and \AUC's correlation with LLM Arena is wider on technical queries (e.g., Math, Expert) versus those unrelated to \mcqa{} (e.g., Creative Writing).

\subsection{Capturing Cross-Scheme Contradictions} \label{subsection:consistency}


We now show alternative scoring schemes' value~in tracking logical consistency \cite{hase-etal-2023-methods}:~a model should not contradict its belief of the answer across schemes \cite{chern2024behonest}.
We test consistency in two ways: 1)~the~answer a model picks in \Norm \underline{must not} be eliminated in \Elim; and 2) \Norm's answer \underline{must} have the most weight in \PPA (or tied).

Figure \ref{fig:mmlu_top_consistency} 
shows that for \mm{}s of the same family, scaling parameters or adding test-time reasoning improves consistency, aligning with prior work \cite{truong-etal-2023-language}. 
Thus, running diverse scoring schemes for the same task like \mcqa{} is a simple way to evaluate which interventions (e.g., scaling size or inference) can improve model consistency.

\subsection{Scoring Schemes Reveal Model Behaviors} \label{subsection:behavior}

While \Norm mainly provides just one accuracy score to study models, alternative schemes also give finer-grained signals into model behavior, like abstention rates, confidence, and adaptability (Figure~\ref{fig:behavior}).
Analyzing this yields interesting trends: 1) GPT~models rarely abstain and state high confidence; 2) weak GPT models struggle to change answers after being told they are incorrect, while stronger GPT models readily do so; 3) weaker open-weight models like Command-R+ and Llama often abstain or hesitate to eliminate choices; and 4) All \mm{}s state high confidence in \CBM and rarely abstain in \NM on easier datasets like ARC and MMLU, but not Super GPQA, showing some awareness of \mcq{} difficulty \cite{guo2017calibration}.
Only evaluating \mcqa{} via \Norm scoring obscures these features of \mm{} behavior.

%% file: figures/correlation.tex
\begin{figure}[t]
    \centering
    \includegraphics[width=\linewidth]{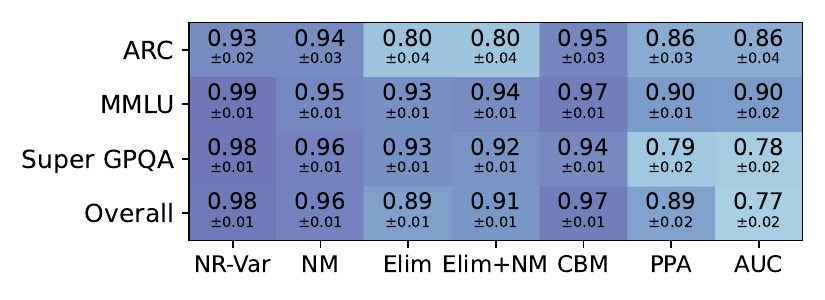}
    \vspace{-4ex}
    \caption{\label{fig:rankings} Spearman correlations between number right~(\Norm) scoring with bootstrapped standard error $(n=1000)$. ``Overall'' micro-averages scores across all datasets. All shift ranks beyond the prompt sensitivity control (\NormVar)---especially \Elim \ElimNM, \PPA and \AUC---so they test diverse model abilities.
    Appendix~\ref{appendix:ranks} has correlations across all schemes.}
\end{figure}

%% file: figures/llm_arena.tex
\begin{figure}[t]
    \centering
    \includegraphics[width=\linewidth]{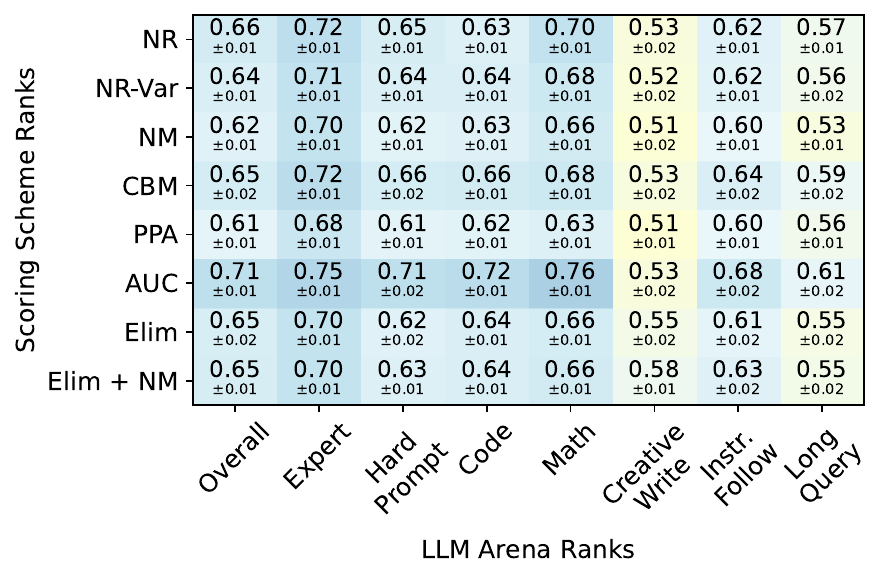}
    \vspace{-4ex}
    \caption{\label{fig:llm_arena} \small Spearman's correlation between \mm{} ranks from scoring schemes (dataset micro-average) versus user preferences of varied query types in LLM Arena. Answering until correct (\AUC) scoring often best predicts model ranks in LLM Arena. Appendix~\ref{appendix:llm_arena} decomposes results across each dataset.}
\end{figure}

%% file: figures/consistency.tex
\begin{figure}[t]
    \centering
    \includegraphics[width=\linewidth]{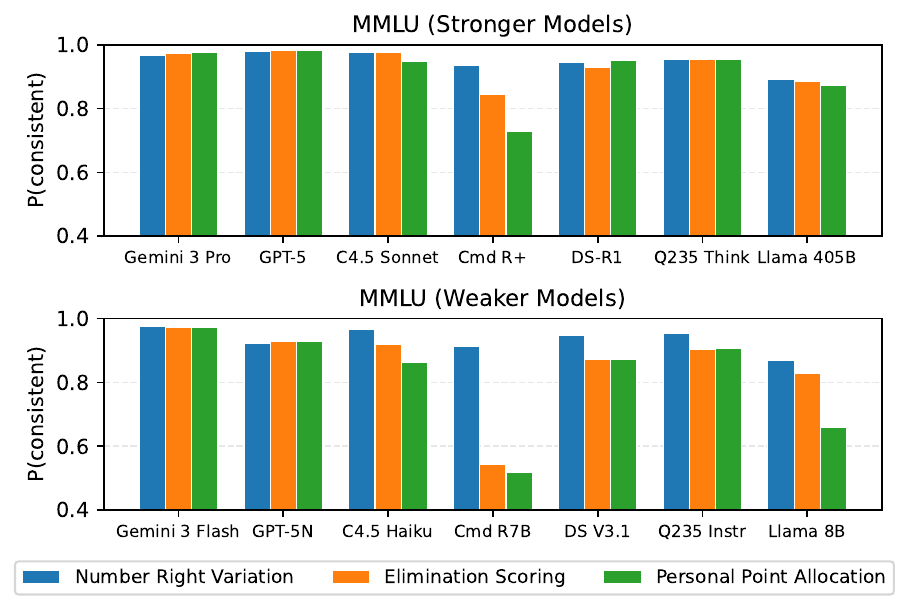}
    \vspace{-3ex}
    \caption{\label{fig:mmlu_top_consistency} \small Logical consistency of scoring schemes with \Norm on MMLU. Weaker models (smaller, no reasoning) tend to be less consistent over schemes. Appendix~\ref{appendix:consistency} has all datasets.}
\end{figure}

%% file: figures/behaviors.tex
\begin{figure*}[t]
    \centering
    \includegraphics[width=\linewidth]{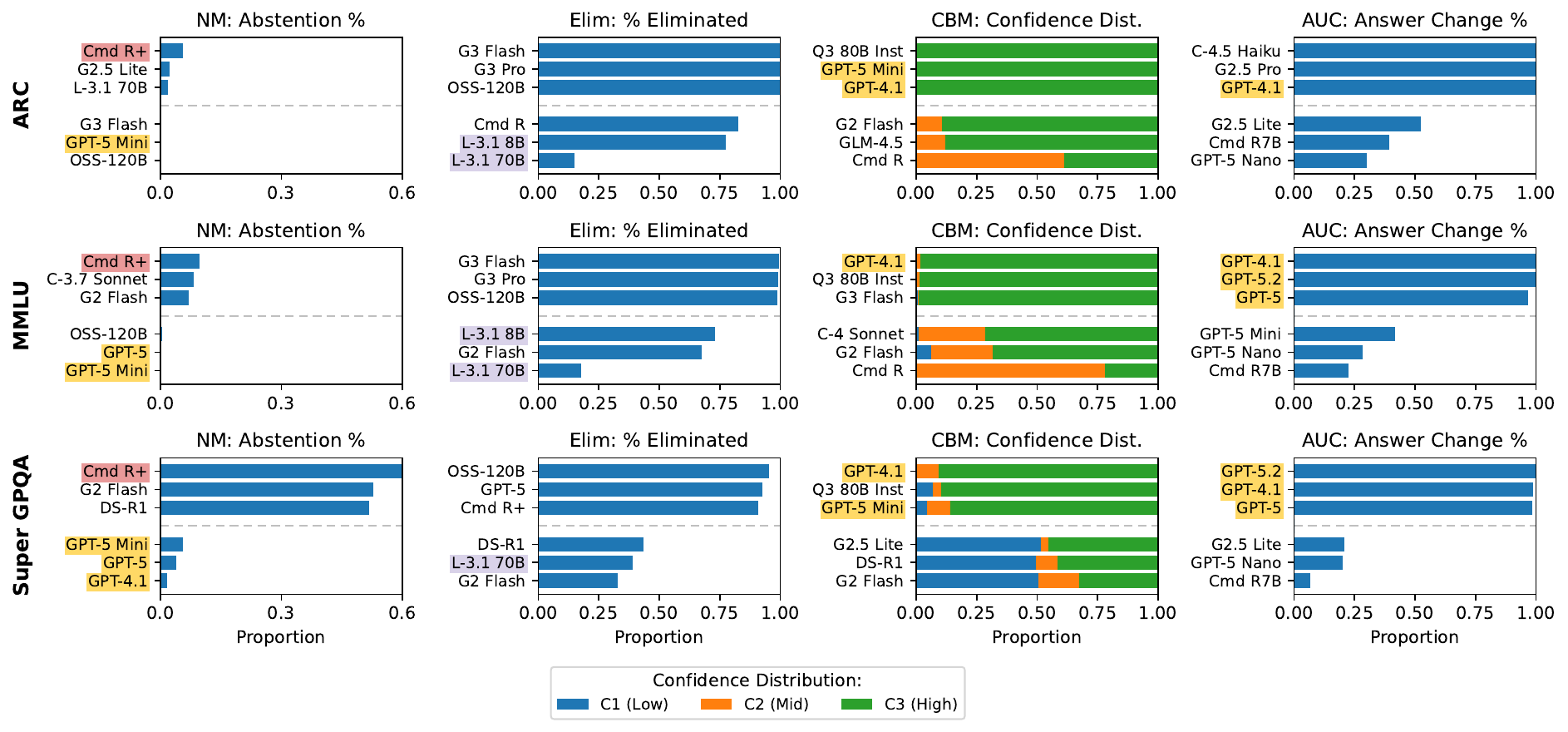}
    \vspace{-3ex}
    \caption{\label{fig:behavior} \small Analysis of abstention rates, option elimination rate, stated confidence, and answer change rates, surfaced via \NM, \Elim, \CBM, and \AUC scoring schemes, respectively. We display the three highest and lowest scoring models under each facet for brevity. These schemes reveal model-specific behaviors: \colorbox{yellowcolor}{\strut stronger GPT models} are confident, rarely abstain, and quickly change answers, \colorbox{redcolor}{\strut Command-R+} often abstains, and \colorbox{purplecolor}{\strut LLaMA-3.1} hesitates to eliminate. Appendix~\ref{appendix:behavior} has all models and trends.}
\end{figure*}

%% file: 2026_arr_mcqa_scoring/sections/50_related_work.tex
\section{Related Work} \label{section:related_work}

\textbf{Scoring Schemes:} \label{subsection:rw_scoring} \mcqa{} exams originally~tested students via number right scoring \cite{kelly1916kansas}, but educators found limitations: it encourages~students to guess when unsure \cite{guilford1936determination}~and cannot discern partial knowledge if students~answer identically \cite{lau2011guessing}.
Educators have thus designed alternatives schemes to improve~abilities \mcqa{} measures \cite{cronbach1955construct}, so we~extend them and assess their benefits in \nlp{}.


\noindent \textbf{\mcqa{} Evaluation:} \label{subsection:rw_evaluation} \mcqa{} is a keystone of~\nlp{} evaluation \cite{reddy1988foundations} as it is simple and mirrors student testing \cite{clark2020f}.
Given~this dominance \cite{liang2023holistic}, work improves the task via open-ended formats \cite{myrzakhan2024open, chandak2025answer}, shortcut reduction \cite{balepur2024your, balepur-etal-2026-test}, and quality control \cite{moore2024automatic, Palta2024PlausiblyPQ, balepur-etal-2026-benchmarker}.
Some~of~our \mcqa schemes have been studied individually: \citet{ma2023poe} and \citet{balepur-etal-2024-easy} test process of elimination, \citet{elhady-etal-2025-wicked} add an ``I don't know'' option for abstention, and \citet{huang2023large} assess self-correction.
We instead compare schemes from education to study rank~shifts, preference agreement, and behaviors. 
Work also studies how scoring can better adapt training and evaluation curricula \citep[e.g., Item Response Theory]{hofmann2025fluid, 10.1162/COLI.a.584}, but our scoring schemes focus on evaluating individual examples.

%% file: 2026_arr_mcqa_scoring/sections/60_conclusion.tex
\section{Conclusion: Scheming Beyond \mcqa{}} \label{section:conclusion}





\nlp{} often critiques \mcqa{} since models can guess \cite{10.1145/3596490, Balepur2024ArtifactsOA} and the task poorly reflects downstream user needs \cite{saxon2024benchmarks, balepur-etal-2025-best}, but it persists due to its simplicity. 
Educators face~similar issues when testing students, so they design~alternative~scoring schemes. 
\nlp{} can take~the same path: easy fixes to prompts and metrics can curb guessing by design, reveal distinct model behaviors, and better predict user preferences, all without changing task~inputs.


More broadly, we argue for human assessments to inform \mm{} scoring scheme design.
In \mcqa{}, we use educational testing, which also applies to programming~\cite{white1993holistic} and writing \cite{tew2010developing} which have well-researched scoring rubrics and assignment creation protocols for students \cite{10.1145/563517.563426}.
For \nlp{} tasks sans human analogues \cite[e.g., deep research]{bragg2025astabench, balepur-etal-2026-language}, a promising path is to study how humans~finish them and which behaviors predict success.
Lastly, the benefits of \AUC's multi-turn setting with simulated user feedback suggests that more classic \nlp{} tasks (e.g., natural language inference, summarization) could be run as multi-turn dialogues, forming user-inspired evaluations \cite{pan-etal-2025-benchmarks, laban2026llms}.
Overall, researchers must treat scoring schemes as a key part of benchmark construction, just like educators~do.

%% file: 2026_arr_mcqa_scoring/sections/70_limitation_ethics.tex


\section{Limitations} \label{section:limitations}

\mm{}s are sensitive to prompting formats, especially in \mcqa{} \cite{pezeshkpour2023large}, so altering prompts could shift model ranks further \cite{alzahrani2024benchmarks}.
To design prompts fairly, we build off the \mcqa{} prompts used in InspectAI \cite{inspect_ai_framework}: a standard evaluation framework from the United Kingdom AI Security Institute.
This also applies to our analysis with LLM Arena \cite{zheng2023judging}, as the platform does not disclose the prompts or inference parameters used, so different configurations could alter our results.
Prompt sensitivity is another interesting aspect of \mm{} evaluation that future work can explore when considering alternative scoring schemes.

Second, while Answer Until Correct (\AUC) is the scoring strategy with the highest agreement with LLM Arena, it is the most expensive, as it requires many turns (cost in Table~\ref{table:cost}).
While still cheaper than crowdsourcing, future work could test alternate ways to reap the benefits of \AUC with less attempts: in Appendix~\ref{appendix:llm_arena}, we show prompting \mm{}s to self-refine their answers just once improves agreement with LLM Arena.
However, our goal is not to argue for a particular \mcqa{} scoring scheme, but to motivate the benefits of evaluating with different scoring schemes, regardless of task.


\section{Ethical Considerations}

Our goal of testing different multiple-choice scoring regimes is to improve evaluation validity, but we never intend to argue that such evaluations are sufficient as pre-deployment checks.
Models that excel in our schemes may still be prone to biases, hallucinations, and producing harmful outputs.

Generative AI (GenAI) was used in this project.
We used Cursor\footnote{https://cursor.com/agents} to design plots, refactor code, and debug errors, and we used ChatGPT to refine paper writing for brevity.
AI tools did not directly write any parts of this paper.
We take full responsibility for any GenAI errors.
By discussing AI usage here, we encourage other researchers to do the same.

\section*{Acknowledgments}

We would like to thank the \abr{clip} lab at the University of Maryland for their support.
In particular, we thank Yapei Chang, Jane Oh, and Ankit Aggarwal for discussions on earlier versions of this paper draft.
This material is based upon work supported by the National Science Foundation under \abr{iis}-2339746 (Rudinger), \abr{iis}-2403436 (Boyd-Graber), and \abr{dge}-2236417 (Balepur).
Boyd-Graber's research is also supported by a gift from Adobe Corporation.
Any opinions, findings, and conclusions or recommendations expressed in this material are those of the author(s) and do not necessarily reflect the views of sponsors.
Access to Cohere models (Command-R, Command-R Plus)~was made possible via a Cohere for AI Research Grant.

%% file: 2026_arr_mcqa_scoring/sections/100_appendix.tex
\section{Appendix} \label{section:appendix}

\subsection{Dataset Details} \label{appendix:dataset}

We select $1000$ random examples from the testing splits of ARC \cite{clark2018think}, MMLU \cite{hendrycks2020measuring}, and Super GPQA \cite{du2025supergpqa}.
All datasets are publicly available, so our experiments are within their intended use.
We did not collect any datasets, so we did not check for PII.
To our knowledge, all questions are in English.

\subsection{Experiment Details} \label{appendix:experiment}

We implement all \mm{}s in InspectAI via LiteLLM\footnote{https://docs.litellm.ai/}.
We access all closed-source \mm{}s via their native APIs (e.g. the OpenAI API for GPT-5).
For open-weight \mm{}s, we use the Cohere API\footnote{https://cohere.com/} for Cohere models and TogetherAI\footnote{https://www.together.ai/} for the rest.
We allocate 24 CPU hours for each experiment.
All results are reported from one run.
The model endpoints are:
\begin{enumerate}[noitemsep]
    \item anthropic/claude-3-7-sonnet-20250219
    \item anthropic/claude-haiku-4-5-20251001
    \item anthropic/claude-sonnet-4-20250514
    \item anthropic/claude-sonnet-4-5-20250929

    \item openai/gpt-4.1-2025-04-14
    \item openai/gpt-5-2025-08-07
    \item openai/gpt-5.2-2025-12-11
    \item openai/gpt-5-mini-2025-08-07
    \item openai/gpt-5-nano-2025-08-07

    \item google/gemini-2.0-flash
    \item google/gemini-2.5-flash-lite
    \item google/gemini-2.5-pro
    \item google/gemini-3-flash-preview
    \item google/gemini-3-pro-preview

    \item openai-api/cohere/command-r-08-2024
    \item openai-api/cohere/command-r7b-12-2024
    \item openai-api/cohere/command-r-plus-08-2024

    \item together/Qwen/Qwen3-235B-A22B-Instruct-2507-tput
    \item together/Qwen/Qwen3-235B-A22B-Thinking-2507
    \item together/Qwen/Qwen3-Next-80B-A3B-Instruct
    \item together/Qwen/Qwen3-Next-80B-A3B-Thinking

    \item together/meta-llama/Meta-Llama-3.1-405B-Instruct-Turbo
    \item together/meta-llama/Meta-Llama-3.1-70B-Instruct-Turbo
    \item together/meta-llama/Meta-Llama-3.1-8B-Instruct-Turbo

    \item together/deepseek-ai/DeepSeek-R1
    \item together/deepseek-ai/DeepSeek-V3.1

    \item together/openai/gpt-oss-120b
    \item together/openai/gpt-oss-20b

    \item together/moonshotai/Kimi-K2-Thinking
    \item together/zai-org/GLM-4.5-Air-FP8
    \item together/zai-org/GLM-4.6
\end{enumerate}

\subsection{Full Ranking Analysis} \label{appendix:ranks}

Tables~\ref{tab:ranks_arc}, \ref{tab:ranks_mmlu}, and \ref{tab:ranks_super_gpqa} have the rankings of all 31 \mm{}s for ARC, MMLU, and SuperGPQA, respectively, while Tables~\ref{tab:scores_arc}, \ref{tab:scores_mmlu}, and \ref{tab:scores_super_gpqa} have the \mm{} scores of these rankings, respectively.
To break ties in ranks, we use Standard Competition or ``1224'' ranking,\footnote{\url{https://en.wikipedia.org/wiki/Ranking\#Standard\_competition\_ranking\_("1224"\_ranking)}} where tied models share the same rank and ensuing positions are skipped accordingly (e.g., 1, 2, 2, 4).
GPT-5 is consistently the strongest across datasets and scoring regimes---ranking the highest overall.

On SuperGPQA, Command-R+ and the GPT-OSS models timed out after 24 hours under \AUC scoring.
Based on their current progress, we estimated that these models would need up to a week to finish, so we did not re-run them, and omitted them from any analysis that involved SuperGPQA.

We also evaluate Pearson's correlation between scores directly (instead of ranks) in Figure~\ref{fig:strategy_correlation_scores} which shows the same trend as \cref{subsection:ranks}: schemes like \Elim, \PPA, and \AUC reward behaviors distinct from \Norm.

\subsection{Full LLM Arena Analysis} \label{appendix:llm_arena}

Figure~\ref{fig:llm_arena_correlation_all} shows the correlation between rankings for each scoring strategy and prompt type in LLM Arena across all datasets.
The trend in \cref{subsection:preferences} holds over each dataset: the Answer Until Correct regime best correlates with preferences from LLM Arena.

To understand the higher agreement from LLM Arena further, we design an extension of \AUC that executes two steps: 1) the model makes an initial prediction; and 2) the model reflects on its answer and decide whether it was correct or incorrect---following self-refine \cite{madaan2023self}. 
We run this scheme on ARC with GPT and open-weight models in \cref{subsection:model}.
The self-refine ablation reaches an agreement of $0.89$, \AUC reaches $0.86$, and \Norm only reaches $0.79$---further indicating that the ability to critically self-reflect is the behavior that correlates well with human preferences.

\subsection{Full Consistency Analysis} \label{appendix:consistency}

Figure~\ref{fig:arc_top_consistency} and \ref{fig:gpqa_top_consistency} replicate Figure~\ref{fig:mmlu_top_consistency} but on ARC and Super GPQA which show a similar trend: smaller models without reasoning tend to be more inconsitent compared to their mode capable counterparts, except for select, strong fronteir models like GPT and Gemini.
Figure~\ref{fig:arc_closed_source}, \ref{fig:mmlu_closed_source}, and \ref{fig:super_gpqa_closed_source} have the consistency analysis results for all scoring strategies in \cref{subsection:consistency} replicated across ARC, MMLU, and Super GPQA for closed-source models, respectively, while Figure~\ref{fig:arc_open_weight}, \ref{fig:mmlu_open_weight}, \ref{fig:super_gpqa_open_weight} show consistency analyses for the same datasets and open-weight models, respectively. 
While closed-source and larger models are fairly consistent between scoring strategies, smaller and open-weight models lag further behind.

\subsection{Full Model Behavior Analysis} \label{appendix:behavior}

Figures~\ref{fig:arc_behavior}, \ref{fig:mmlu_behavior}, and \ref{fig:super_gpqa_behavior} show the behavioral analysis from \cref{subsection:behavior} across all datasets, models, and evaluated scoring regimes.
Tables~\ref{tab:scores_arc}, \ref{tab:scores_mmlu}, and \ref{tab:scores_super_gpqa} have the \mm{} scores of these rankings, respectively.

\paragraph{GPT models rarely abstain and state high confidence.}
Ranking models by \NM abstention rate in Figures~\ref{fig:arc_behavior} and~\ref{fig:mmlu_behavior}, GPT models occupy six of the ten lowest-abstaining positions; on the harder Super GPQA (Figure~\ref{fig:super_gpqa_behavior}), six of the eight least-abstaining models are GPT variants.
An exception is GPT-OSS-20B, a weaker open-sourced GPT model that falls near the upper half of the abstention ranking across all datasets.
GPT models state high confidence as well:
while most \mm{}s avoid low-confidence responses on easier datasets like ARC and MMLU, GPT models take the six positions with the fewest low-confidence responses on the harder Super GPQA (Figure~\ref{fig:super_gpqa_behavior}).

\paragraph{Stronger GPT models readily revise answers after feedback; weaker ones struggle to do so.}
The answer-change behavior from the figures supports this directly: GPT-4.1, GPT-5, GPT-5.2, and GPT-OSS-120B fill four of the top six answer-change positions on ARC (Figure~\ref{fig:arc_behavior}) and all four of the top four on MMLU (Figure~\ref{fig:mmlu_behavior}).
On Super GPQA, GPT-OSS-120B exceeded our time budget, yet the other three stronger GPT models still claim the top three positions (Figure~\ref{fig:super_gpqa_behavior}).
By contrast, GPT-5~Nano finishes last on ARC and second-to-last on Super GPQA in answer-change rate, and GPT-5~Mini ranks near the bottom on ARC and MMLU---the smallest GPT models are far less willing to revise answers upon feedback.

\paragraph{Weaker open-weight models often abstain or hesitate to eliminate choices.}
Command-R+ holds the single highest \NM abstention rate on all three datasets (Figures~\ref{fig:arc_behavior}--\ref{fig:super_gpqa_behavior}), consistently the most likely model to pass rather than commit to an answer.
Llama models show the same avoidance tendency, but in \Elim rather than abstention: across all three datasets, Llama~3.1~70B ranks last (31st on ARC and MMLU, 29th on Super GPQA), Llama~3.1~8B remains near the bottom (29th on ARC and MMLU, 28th on Super GPQA), and even Llama~3.1~405B falls in the lower quarter (24th--25th on ARC and MMLU, 23rd on Super GPQA)---indicating a systematic reluctance to rule out incorrect options for Llama models.

\paragraph{All \mm{}s show high confidence and low abstention on easier datasets, but not on Super GPQA.}
On ARC and MMLU, most models predominantly express high confidence and rarely abstain.
On Super GPQA this changes markedly: abstention rates rise and models that predominantly expressed high confidence on easier tasks shift toward low and mid confidence (Figures~\ref{fig:arc_behavior}--\ref{fig:super_gpqa_behavior}).
Even the \Elim elimination rates decline with difficulty, as models eliminate fewer choices on harder questions.


\subsection{Prompts} \label{appendix:prompt}

We show the prompts for all eight scoring schemes in Prompts~\ref{prompt:normal}--\ref{prompt:auc}.
Each scheme adapts the base \Norm{} prompt (Prompt~\ref{prompt:normal})---initially drawn from InspectAI \cite{inspect_ai_framework}---to alter the \textit{response mode}: how the \mm{} responds to the \mcq{}. Afterwards, the scheme may also adjust the \textit{scoring rule} via custom metrics.

\paragraph{Number Right (\Norm, Prompt~\ref{prompt:normal}).}
This is the standard prompt used in \mcqa{}: the response mode asks the model to select one answer letter, and the scoring rule rewards only correct answers \citep{kelly1916kansas}.

\paragraph{Control Variation (\NormVar, Prompt~\ref{prompt:variation}).}
We rephrase \Norm{}'s task {\it instruction} with synonymous language while leaving both the response mode and scoring rule unchanged, to isolate sensitivity to prompt wording \citep{alzahrani2024benchmarks}.

\paragraph{Negative Marking (\NM, Prompt~\ref{prompt:negative_marking}).}
We adapt \Norm{}'s response mode to let \mm{}s abstain (via \texttt{ABSTAIN} in the answer field), and update the scoring rule to penalize wrong answers while rewarding abstaining over guessing \citep{holt2006analysis}.

\paragraph{Elimination (\Elim, Prompt~\ref{prompt:elimination}).}
Instead of selecting a correct answer, \mm{}s output a list of choices they believe are wrong.
The scoring rule is updated to grant partial credit for each eliminated distractor, but revokes all credit if the correct answer is eliminated \citep{coombs1956assessment}.

\paragraph{Elimination with Negative Marking (\ElimNM, Prompt~\ref{prompt:elimination_negative}).}
We keep \Elim{}'s elimination response mode but adapt the scoring rule to additionally penalize eliminating the correct answer, combining partial-knowledge credit with a guessing deterrent \citep{coombs1956assessment}.

\paragraph{Confidence-Based Marking (\CBM, Prompt~\ref{prompt:confidence}).}
We extend \Norm{}'s response mode to require a discrete confidence level alongside the answer, and update the scoring rule so that higher declared confidence amplifies both the reward for correct answers and the penalty for wrong ones \citep{gardner2006confidence}.

\paragraph{Personal Point Allocation (\PPA, Prompt~\ref{prompt:personal_point}).}
We replace \Norm{}'s single-answer response mode with a probability distribution over all choices, and replace the binary scoring rule with the Brier score to reward both accuracy and calibration \citep{macneil2023examining}.

\paragraph{Answer Until Correct (\AUC, Prompt~\ref{prompt:auc}).}
We adapt \Norm{}'s response mode to allow multiple attempts, injecting a feedback block after each wrong answer that lists the \mm{}'s prior incorrect choices, and update the scoring rule to reward answering correctly in fewer attempts \citep{wilcox1982some}.

\clearpage

\input{appendix/robustness}
\input{appendix/cost}
\input{appendix/ranks}
\input{appendix/strategy_correlation}
\input{appendix/llm_arena_all}
\input{appendix/consistency}
\input{appendix/behavior}
\clearpage
\input{appendix/prompts}

%% file: appendix/robustness.tex
\begin{table*}[t]
    \centering
    \begin{tabular}{lc}
        \toprule
        \textbf{Scheme} & \textbf{Agreement with Number Right} \\
        \midrule
        Rephrased Prompt & 0.93 \\
        Shuffle Choices           & 0.96 \\
        Numbers as Choices        & 0.98 \\
        \NM                        & 0.94 \\
        \Elim                      & 0.80 \\
        \ElimNM                   & 0.80 \\
        \CBM                       & 0.95 \\
        \PPA                       & 0.86 \\
        \AUC                       & 0.86 \\
        \bottomrule
    \end{tabular}
    \caption{Agreement between alternative scoring schemes with Number Right on ARC, contextualized by three prompt variation baselines for Number Right: rephrasing the text of the instructions, shuffling answer choices, and replacing letters with numbers in the choices.}
    \label{table:robustness}
\end{table*}

%% file: appendix/cost.tex
\begin{table*}[t]
    \small
    \centering
    \begin{tabular}{lrrrrrr}
        \toprule
        \textbf{Scoring Scheme}
        & \textbf{ARC (In)}
        & \textbf{ARC (Out)}
        & \textbf{MMLU (In)}
        & \textbf{MMLU (Out)}
        & \textbf{SuperGPQA (In)}
        & \textbf{SuperGPQA (Out)} \\
        \midrule
        \Norm      & 207 & 122 & 277   & 186 & 1,602  & 1,236 \\
        \NormVar  & 206 & 101 & 271   & 151 & 1,297  & 923   \\
        \NM      & 260 & 140 & 326   & 197 & 1,389  & 791   \\
        \AUC     & 492 & 168 & 1,363 & 436 & 45,695 & 4,088 \\
        \CBM     & 384 & 153 & 458   & 210 & 1,647  & 1,142 \\
        \PPA     & 428 & 211 & 492   & 276 & 1,448  & 1,076 \\
        \Elim    & 298 & 179 & 373   & 259 & 1,489  & 980   \\
        \ElimNM & 297 & 184 & 379   & 278 & 1,462  & 984   \\
        \bottomrule
    \end{tabular}
    \caption{The median number of input and output tokens generated on each dataset across the distribution of models tested. While \AUC benefits \mcqa{} evaluation, we caveat that the scheme's multi-turn design increases evaluation costs---especially on difficult datasets like SuperGPQA with many choices.}
    \label{table:cost}
\end{table*}

%% file: appendix/ranks.tex
\input{appendix/ARC_rank}
\input{appendix/MMLU_rank}
\input{appendix/GPQA_rank}

%% file: appendix/ARC_rank.tex
\begin{table*}
\centering
\small
\renewcommand{\arraystretch}{0.9}
\begin{tabular}{lccccccccc}
\toprule
Model & \Norm & \NormVar & \NM & \AUC & \CBM & \PPA & \Elim & \ElimNM & Overall \\
\midrule
Claude 3.7 Sonnet & 12 & 15 & 22 & 25 & 18 & 19 & 25 & 25 & 23 \\
Claude Haiku 4.5 & 23 & 22 & 19 & 19 & 22 & 21 & 21 & 20 & 12 \\
Claude Sonnet 4 & 1 & 4 & 5 & 13 & 3 & 10 & 13 & 12 & 11 \\
Claude Sonnet 4.5 & 1 & 2 & 1 & 1 & 4 & 5 & 18 & 18 & 10 \\
Gemini 2.0 Flash & 20 & 21 & 21 & 18 & 21 & 23 & 26 & 26 & 25 \\
Gemini 2.5 Flash Lite & 25 & 25 & 25 & 26 & 26 & 24 & 23 & 24 & 26 \\
Gemini 2.5 Pro & 7 & 7 & 3 & 3 & 1 & 4 & 3 & 3 & 4 \\
Gemini 3 Flash & 3 & 1 & 2 & 4 & 9 & 1 & 1 & 1 & 2 \\
Gemini 3 Pro & 3 & 4 & 6 & 5 & 5 & 2 & 2 & 2 & 6 \\
Command R & 28 & 28 & 29 & 29 & 29 & 28 & 28 & 28 & 28 \\
Command R+ & 26 & 27 & 28 & 28 & 28 & 25 & 27 & 27 & 27 \\
Command R7B & 31 & 29 & 31 & 31 & 31 & 29 & 30 & 31 & 30 \\
GPT-4.1 & 17 & 8 & 15 & 7 & 15 & 11 & 11 & 9 & 16 \\
GPT-5 & 3 & 2 & 4 & 2 & 2 & 3 & 3 & 3 & 1 \\
GPT-5 Mini & 6 & 6 & 7 & 16 & 6 & 6 & 7 & 5 & 3 \\
GPT-5 Nano & 20 & 23 & 17 & 20 & 19 & 18 & 15 & 15 & 8 \\
GPT-5.2 & 8 & 9 & 9 & 6 & 8 & 8 & 9 & 13 & 13 \\
Qwen3-235B Instruct & 16 & 17 & 16 & 11 & 13 & 20 & 22 & 22 & 14 \\
Qwen3-235B Thinking & 12 & 12 & 8 & 8 & 12 & 13 & 8 & 7 & 9 \\
Qwen3-80B Instruct & 18 & 19 & 18 & 13 & 16 & 22 & 19 & 19 & 17 \\
Qwen3-80B Thinking & 15 & 16 & 13 & 17 & 14 & 15 & 12 & 10 & 15 \\
DeepSeek-R1 & 10 & 9 & 13 & 13 & 10 & 12 & 9 & 8 & 18 \\
DeepSeek-V3.1 & 19 & 17 & 24 & 23 & 19 & 17 & 20 & 21 & 22 \\
Llama 3.1 405B & 28 & 26 & 20 & 24 & 24 & 26 & 24 & 23 & 24 \\
Llama 3.1 70B & 27 & 31 & 27 & 27 & 27 & 31 & 31 & 30 & 31 \\
Llama 3.1 8B & 30 & 30 & 30 & 30 & 30 & 27 & 29 & 29 & 29 \\
Kimi-K2 Thinking & 9 & 9 & 12 & 10 & 11 & 7 & 5 & 6 & 7 \\
GPT-OSS-120B & 10 & 19 & 10 & 9 & 17 & 9 & 6 & 11 & 5 \\
GPT-OSS-20B & 24 & 24 & 26 & 22 & 25 & 14 & 14 & 17 & 19 \\
GLM-4.5 Air & 22 & 12 & 23 & 21 & 23 & 16 & 16 & 14 & 20 \\
GLM-4.6 & 14 & 14 & 10 & 12 & 7 & 30 & 17 & 16 & 21 \\
\bottomrule
\end{tabular}
\caption{Model ranks across scoring schemes for ARC}
\label{tab:ranks_arc}
\end{table*}

\begin{table*}
\centering
\small
\renewcommand{\arraystretch}{0.9}
\begin{tabular}{lccccccccc}
\toprule
Model & \Norm & \NormVar & \NM & \AUC & \CBM & \PPA & \Elim & \ElimNM & Overall \\
\midrule
Claude 3.7 Sonnet & 0.967 & 0.963 & 0.938 & 0.95 & 0.958 & 0.943 & 0.851 & 0.791 & 0.681 \\
Claude Haiku 4.5 & 0.952 & 0.956 & 0.941 & 0.979 & 0.954 & 0.931 & 0.897 & 0.834 & 0.76 \\
Claude Sonnet 4 & 0.981 & 0.977 & 0.968 & 0.981 & 0.977 & 0.966 & 0.944 & 0.927 & 0.768 \\
Claude Sonnet 4.5 & 0.981 & 0.978 & 0.975 & 0.991 & 0.976 & 0.973 & 0.904 & 0.881 & 0.772 \\
Gemini 2.0 Flash & 0.956 & 0.957 & 0.938 & 0.979 & 0.956 & 0.922 & 0.837 & 0.788 & 0.646 \\
Gemini 2.5 Flash Lite & 0.94 & 0.942 & 0.921 & 0.946 & 0.944 & 0.917 & 0.86 & 0.811 & 0.623 \\
Gemini 2.5 Pro & 0.975 & 0.971 & 0.971 & 0.99 & 0.981 & 0.974 & 0.973 & 0.959 & 0.832 \\
Gemini 3 Flash & 0.98 & 0.983 & 0.975 & 0.989 & 0.97 & 0.985 & 0.982 & 0.971 & 0.872 \\
Gemini 3 Pro & 0.98 & 0.977 & 0.968 & 0.987 & 0.976 & 0.983 & 0.98 & 0.965 & 0.801 \\
Command R & 0.863 & 0.862 & 0.797 & 0.914 & 0.856 & 0.716 & 0.702 & 0.593 & 0.552 \\
Command R+ & 0.89 & 0.888 & 0.822 & 0.938 & 0.889 & 0.829 & 0.758 & 0.651 & 0.591 \\
Command R7B & 0.8 & 0.807 & 0.729 & 0.842 & 0.794 & 0.657 & 0.544 & 0.318 & 0.462 \\
GPT-4.1 & 0.963 & 0.968 & 0.946 & 0.986 & 0.962 & 0.96 & 0.95 & 0.934 & 0.752 \\
GPT-5 & 0.98 & 0.978 & 0.97 & 0.99 & 0.979 & 0.979 & 0.973 & 0.959 & 0.896 \\
GPT-5 Mini & 0.976 & 0.975 & 0.962 & 0.981 & 0.974 & 0.972 & 0.962 & 0.945 & 0.852 \\
GPT-5 Nano & 0.956 & 0.954 & 0.945 & 0.967 & 0.957 & 0.947 & 0.937 & 0.893 & 0.798 \\
GPT-5.2 & 0.972 & 0.967 & 0.954 & 0.986 & 0.972 & 0.969 & 0.959 & 0.926 & 0.759 \\
Qwen3-235B Instruct & 0.964 & 0.961 & 0.945 & 0.983 & 0.965 & 0.937 & 0.887 & 0.821 & 0.758 \\
Qwen3-235B Thinking & 0.967 & 0.966 & 0.955 & 0.985 & 0.967 & 0.959 & 0.961 & 0.937 & 0.795 \\
Qwen3-80B Instruct & 0.962 & 0.96 & 0.943 & 0.981 & 0.961 & 0.931 & 0.901 & 0.844 & 0.743 \\
Qwen3-80B Thinking & 0.965 & 0.962 & 0.95 & 0.98 & 0.963 & 0.95 & 0.948 & 0.933 & 0.755 \\
DeepSeek-R1 & 0.968 & 0.967 & 0.95 & 0.981 & 0.968 & 0.96 & 0.959 & 0.935 & 0.727 \\
DeepSeek-V3.1 & 0.96 & 0.961 & 0.921 & 0.96 & 0.957 & 0.948 & 0.898 & 0.833 & 0.7 \\
Llama 3.1 405B & 0.863 & 0.914 & 0.94 & 0.955 & 0.95 & 0.809 & 0.857 & 0.819 & 0.647 \\
Llama 3.1 70B & 0.885 & 0.768 & 0.908 & 0.943 & 0.922 & 0 & 0.142 & 0.421 & 0.459 \\
Llama 3.1 8B & 0.804 & 0.804 & 0.756 & 0.894 & 0.816 & 0.748 & 0.684 & 0.56 & 0.52 \\
Kimi-K2 Thinking & 0.97 & 0.967 & 0.953 & 0.984 & 0.967 & 0.97 & 0.967 & 0.943 & 0.799 \\
GPT-OSS-120B & 0.968 & 0.96 & 0.953 & 0.984 & 0.96 & 0.969 & 0.963 & 0.929 & 0.824 \\
GPT-OSS-20B & 0.943 & 0.947 & 0.909 & 0.966 & 0.945 & 0.951 & 0.941 & 0.886 & 0.718 \\
GLM-4.5 Air & 0.954 & 0.966 & 0.933 & 0.966 & 0.954 & 0.949 & 0.917 & 0.908 & 0.713 \\
GLM-4.6 & 0.966 & 0.965 & 0.953 & 0.982 & 0.972 & 0.483 & 0.904 & 0.888 & 0.706 \\
\bottomrule
\end{tabular}
\caption{Model scores across scoring schemes for ARC}
\label{tab:scores_arc}
\end{table*}

%% file: appendix/MMLU_rank.tex
\begin{table*}
\centering
\small
\renewcommand{\arraystretch}{0.9}
\begin{tabular}{lccccccccc}
\toprule
Model & \Norm & \NormVar & \NM & \AUC & \CBM & \PPA & \Elim & \ElimNM & Overall \\
\midrule
Claude 3.7 Sonnet & 15 & 14 & 24 & 26 & 14 & 15 & 23 & 23 & 23 \\
Claude Haiku 4.5 & 19 & 17 & 15 & 14 & 18 & 21 & 19 & 18 & 12 \\
Claude Sonnet 4 & 8 & 7 & 8 & 17 & 9 & 10 & 14 & 11 & 11 \\
Claude Sonnet 4.5 & 5 & 5 & 6 & 5 & 7 & 6 & 11 & 7 & 10 \\
Gemini 2.0 Flash & 23 & 20 & 22 & 22 & 24 & 24 & 26 & 25 & 25 \\
Gemini 2.5 Flash Lite & 25 & 26 & 27 & 28 & 27 & 23 & 24 & 26 & 26 \\
Gemini 2.5 Pro & 4 & 4 & 5 & 3 & 5 & 4 & 4 & 4 & 4 \\
Gemini 3 Flash & 1 & 2 & 2 & 2 & 2 & 3 & 3 & 2 & 2 \\
Gemini 3 Pro & 3 & 3 & 3 & 4 & 3 & 2 & 1 & 1 & 6 \\
Command R & 29 & 29 & 29 & 29 & 29 & 26 & 28 & 30 & 28 \\
Command R+ & 27 & 27 & 28 & 25 & 28 & 25 & 27 & 27 & 27 \\
Command R7B & 31 & 30 & 31 & 31 & 30 & 29 & 30 & 31 & 30 \\
GPT-4.1 & 16 & 18 & 17 & 9 & 19 & 17 & 14 & 13 & 16 \\
GPT-5 & 2 & 1 & 1 & 1 & 1 & 1 & 2 & 3 & 1 \\
GPT-5 Mini & 7 & 6 & 4 & 11 & 4 & 5 & 6 & 9 & 3 \\
GPT-5 Nano & 14 & 15 & 13 & 20 & 15 & 13 & 12 & 16 & 8 \\
GPT-5.2 & 11 & 11 & 14 & 6 & 11 & 12 & 10 & 14 & 13 \\
Qwen3-235B Instruct & 17 & 19 & 18 & 13 & 16 & 18 & 22 & 22 & 14 \\
Qwen3-235B Thinking & 6 & 9 & 9 & 12 & 8 & 9 & 7 & 10 & 9 \\
Qwen3-80B Instruct & 20 & 21 & 21 & 15 & 23 & 22 & 21 & 21 & 17 \\
Qwen3-80B Thinking & 18 & 16 & 16 & 18 & 17 & 16 & 13 & 12 & 15 \\
DeepSeek-R1 & 12 & 10 & 11 & 10 & 13 & 11 & 9 & 8 & 18 \\
DeepSeek-V3.1 & 21 & 22 & 25 & 23 & 21 & 20 & 20 & 20 & 22 \\
Llama 3.1 405B & 28 & 25 & 19 & 27 & 20 & 27 & 25 & 24 & 24 \\
Llama 3.1 70B & 26 & 28 & 26 & 24 & 26 & 31 & 31 & 28 & 31 \\
Llama 3.1 8B & 30 & 31 & 30 & 30 & 31 & 28 & 29 & 29 & 29 \\
Kimi-K2 Thinking & 10 & 8 & 7 & 7 & 6 & 8 & 5 & 5 & 7 \\
GPT-OSS-120B & 9 & 11 & 10 & 8 & 10 & 7 & 8 & 6 & 5 \\
GPT-OSS-20B & 24 & 24 & 23 & 21 & 25 & 19 & 18 & 19 & 19 \\
GLM-4.5 Air & 22 & 23 & 20 & 19 & 22 & 14 & 17 & 17 & 20 \\
GLM-4.6 & 13 & 13 & 12 & 16 & 12 & 30 & 16 & 15 & 21 \\
\bottomrule
\end{tabular}
\caption{Model ranks across scoring schemes for MMLU}
\label{tab:ranks_mmlu}
\end{table*}

\begin{table*}
\centering
\small
\renewcommand{\arraystretch}{0.9}
\begin{tabular}{lccccccccc}
\toprule
Model & \Norm & \NormVar & \NM & \AUC & \CBM & \PPA & \Elim & \ElimNM & Overall \\
\midrule
Claude 3.7 Sonnet & 0.867 & 0.867 & 0.767 & 0.83 & 0.876 & 0.868 & 0.703 & 0.572 & 0.681 \\
Claude Haiku 4.5 & 0.857 & 0.858 & 0.819 & 0.92 & 0.862 & 0.823 & 0.785 & 0.673 & 0.76 \\
Claude Sonnet 4 & 0.896 & 0.901 & 0.868 & 0.905 & 0.901 & 0.89 & 0.831 & 0.769 & 0.768 \\
Claude Sonnet 4.5 & 0.915 & 0.907 & 0.878 & 0.945 & 0.902 & 0.904 & 0.844 & 0.805 & 0.772 \\
Gemini 2.0 Flash & 0.816 & 0.835 & 0.785 & 0.863 & 0.83 & 0.766 & 0.635 & 0.556 & 0.646 \\
Gemini 2.5 Flash Lite & 0.786 & 0.785 & 0.724 & 0.81 & 0.794 & 0.779 & 0.683 & 0.533 & 0.623 \\
Gemini 2.5 Pro & 0.921 & 0.913 & 0.879 & 0.953 & 0.905 & 0.919 & 0.913 & 0.849 & 0.832 \\
Gemini 3 Flash & 0.94 & 0.935 & 0.91 & 0.959 & 0.936 & 0.933 & 0.925 & 0.885 & 0.872 \\
Gemini 3 Pro & 0.935 & 0.934 & 0.89 & 0.951 & 0.927 & 0.933 & 0.934 & 0.896 & 0.801 \\
Command R & 0.667 & 0.67 & 0.55 & 0.777 & 0.746 & 0.628 & 0.551 & 0.307 & 0.552 \\
Command R+ & 0.751 & 0.756 & 0.618 & 0.833 & 0.759 & 0.699 & 0.607 & 0.398 & 0.591 \\
Command R7B & 0.627 & 0.627 & 0.511 & 0.663 & 0.648 & 0.595 & 0.412 & 0.08 & 0.462 \\
GPT-4.1 & 0.862 & 0.857 & 0.812 & 0.93 & 0.86 & 0.854 & 0.831 & 0.751 & 0.752 \\
GPT-5 & 0.938 & 0.937 & 0.92 & 0.969 & 0.946 & 0.948 & 0.925 & 0.873 & 0.896 \\
GPT-5 Mini & 0.899 & 0.906 & 0.879 & 0.926 & 0.91 & 0.914 & 0.881 & 0.793 & 0.852 \\
GPT-5 Nano & 0.874 & 0.866 & 0.827 & 0.884 & 0.874 & 0.875 & 0.839 & 0.722 & 0.798 \\
GPT-5.2 & 0.88 & 0.878 & 0.825 & 0.939 & 0.89 & 0.882 & 0.845 & 0.747 & 0.759 \\
Qwen3-235B Instruct & 0.861 & 0.853 & 0.81 & 0.925 & 0.866 & 0.85 & 0.731 & 0.587 & 0.758 \\
Qwen3-235B Thinking & 0.903 & 0.888 & 0.863 & 0.925 & 0.902 & 0.9 & 0.876 & 0.788 & 0.795 \\
Qwen3-80B Instruct & 0.844 & 0.834 & 0.789 & 0.915 & 0.834 & 0.804 & 0.744 & 0.599 & 0.743 \\
Qwen3-80B Thinking & 0.858 & 0.86 & 0.814 & 0.897 & 0.865 & 0.857 & 0.838 & 0.763 & 0.755 \\
DeepSeek-R1 & 0.879 & 0.883 & 0.836 & 0.928 & 0.879 & 0.884 & 0.863 & 0.799 & 0.727 \\
DeepSeek-V3.1 & 0.84 & 0.833 & 0.763 & 0.847 & 0.842 & 0.837 & 0.771 & 0.626 & 0.7 \\
Llama 3.1 405B & 0.689 & 0.794 & 0.796 & 0.826 & 0.848 & 0.626 & 0.652 & 0.56 & 0.647 \\
Llama 3.1 70B & 0.754 & 0.688 & 0.736 & 0.842 & 0.808 & 0.002 & 0.142 & 0.319 & 0.459 \\
Llama 3.1 8B & 0.628 & 0.625 & 0.516 & 0.737 & 0.642 & 0.624 & 0.537 & 0.318 & 0.52 \\
Kimi-K2 Thinking & 0.892 & 0.897 & 0.871 & 0.936 & 0.903 & 0.901 & 0.894 & 0.815 & 0.799 \\
GPT-OSS-120B & 0.893 & 0.878 & 0.836 & 0.935 & 0.895 & 0.902 & 0.873 & 0.806 & 0.824 \\
GPT-OSS-20B & 0.806 & 0.825 & 0.772 & 0.868 & 0.827 & 0.845 & 0.793 & 0.668 & 0.718 \\
GLM-4.5 Air & 0.829 & 0.832 & 0.791 & 0.891 & 0.842 & 0.875 & 0.81 & 0.681 & 0.713 \\
GLM-4.6 & 0.875 & 0.872 & 0.831 & 0.91 & 0.889 & 0.461 & 0.818 & 0.726 & 0.706 \\
\bottomrule
\end{tabular}
\caption{Model scores across scoring schemes for MMLU}
\label{tab:scores_mmlu}
\end{table*}

%% file: appendix/GPQA_rank.tex
\begin{table*}
\centering
\small
\renewcommand{\arraystretch}{0.9}
\begin{tabular}{lccccccccc}
\toprule
Model & \Norm & \NormVar & \NM & \AUC & \CBM & \PPA & \Elim & \ElimNM & Overall \\
\midrule
Claude 3.7 Sonnet & 20 & 21 & 24 & 26 & 23 & 16 & 24 & 26 & 23 \\
Claude Haiku 4.5 & 14 & 10 & 9 & 10 & 8 & 13 & 7 & 8 & 12 \\
Claude Sonnet 4 & 11 & 11 & 11 & 19 & 12 & 11 & 14 & 12 & 11 \\
Claude Sonnet 4.5 & 15 & 14 & 12 & 15 & 11 & 9 & 19 & 11 & 10 \\
Gemini 2.0 Flash & 26 & 26 & 25 & 25 & 27 & 28 & 26 & 25 & 25 \\
Gemini 2.5 Flash Lite & 30 & 30 & 27 & 27 & 30 & 30 & 25 & 24 & 26 \\
Gemini 2.5 Pro & 6 & 4 & 4 & 9 & 6 & 7 & 4 & 4 & 4 \\
Gemini 3 Flash & 3 & 3 & 2 & 6 & 3 & 3 & 2 & 2 & 2 \\
Gemini 3 Pro & 11 & 16 & 16 & 17 & 15 & 6 & 8 & 7 & 6 \\
Command R & 29 & 28 & 29 & 21 & 20 & 18 & 30 & 31 & 28 \\
Command R+ & 25 & 25 & 30 & --- & 24 & 29 & 27 & 29 & 27 \\
Command R7B & 31 & 31 & 31 & 28 & 31 & 12 & 31 & 30 & 30 \\
GPT-4.1 & 17 & 19 & 18 & 4 & 22 & 21 & 11 & 14 & 16 \\
GPT-5 & 1 & 1 & 1 & 1 & 1 & 1 & 1 & 1 & 1 \\
GPT-5 Mini & 2 & 2 & 3 & 8 & 2 & 2 & 3 & 3 & 3 \\
GPT-5 Nano & 4 & 5 & 6 & 13 & 5 & 5 & 6 & 5 & 8 \\
GPT-5.2 & 18 & 18 & 17 & 3 & 14 & 15 & 18 & 22 & 13 \\
Qwen3-235B Instruct & 8 & 8 & 13 & 2 & 7 & 8 & 13 & 16 & 14 \\
Qwen3-235B Thinking & 7 & 7 & 7 & 11 & 9 & 14 & 10 & 9 & 9 \\
Qwen3-80B Instruct & 10 & 11 & 15 & 7 & 13 & 10 & 12 & 17 & 17 \\
Qwen3-80B Thinking & 13 & 13 & 10 & 12 & 17 & 22 & 20 & 13 & 15 \\
DeepSeek-R1 & 21 & 22 & 20 & 16 & 26 & 25 & 22 & 20 & 18 \\
DeepSeek-V3.1 & 19 & 17 & 22 & 23 & 19 & 23 & 21 & 21 & 22 \\
Llama 3.1 405B & 24 & 22 & 19 & 18 & 25 & 27 & 23 & 23 & 24 \\
Llama 3.1 70B & 27 & 27 & 26 & 22 & 28 & 31 & 29 & 27 & 31 \\
Llama 3.1 8B & 28 & 29 & 28 & 24 & 29 & 26 & 28 & 28 & 29 \\
Kimi-K2 Thinking & 8 & 9 & 8 & 5 & 10 & 17 & 9 & 10 & 7 \\
GPT-OSS-120B & 5 & 6 & 5 & --- & 4 & 4 & 5 & 6 & 5 \\
GPT-OSS-20B & 22 & 20 & 21 & --- & 21 & 19 & 17 & 18 & 19 \\
GLM-4.5 Air & 23 & 24 & 23 & 20 & 18 & 20 & 16 & 19 & 20 \\
GLM-4.6 & 16 & 15 & 14 & 14 & 16 & 24 & 15 & 15 & 21 \\
\bottomrule
\end{tabular}
\caption{Model ranks across scoring schemes for SuperGPQA. Dashes (---) denote models that could not complete the evaluation run in time, and were thus omitted from analyses.}
\label{tab:ranks_super_gpqa}
\end{table*}

\begin{table*}
\centering
\renewcommand{\arraystretch}{0.9}
\small
\begin{tabular}{lccccccccc}
\toprule
Model & \Norm & \NormVar & \NM & \AUC & \CBM & \PPA & \Elim & \ElimNM & Overall \\
\midrule
Claude 3.7 Sonnet & 0.364 & 0.333 & 0.212 & 0.291 & 0.411 & 0.512 & 0.314 & 0.197 & 0.681 \\
Claude Haiku 4.5 & 0.427 & 0.477 & 0.432 & 0.725 & 0.547 & 0.525 & 0.582 & 0.495 & 0.76 \\
Claude Sonnet 4 & 0.469 & 0.466 & 0.406 & 0.5 & 0.523 & 0.535 & 0.445 & 0.401 & 0.768 \\
Claude Sonnet 4.5 & 0.422 & 0.435 & 0.384 & 0.602 & 0.541 & 0.564 & 0.417 & 0.41 & 0.772 \\
Gemini 2.0 Flash & 0.217 & 0.228 & 0.183 & 0.398 & 0.272 & 0.314 & 0.265 & 0.205 & 0.646 \\
Gemini 2.5 Flash Lite & 0.179 & 0.193 & 0.16 & 0.251 & 0.197 & 0.268 & 0.31 & 0.212 & 0.623 \\
Gemini 2.5 Pro & 0.555 & 0.624 & 0.542 & 0.737 & 0.588 & 0.635 & 0.639 & 0.604 & 0.832 \\
Gemini 3 Flash & 0.673 & 0.66 & 0.651 & 0.775 & 0.716 & 0.717 & 0.76 & 0.728 & 0.872 \\
Gemini 3 Pro & 0.469 & 0.42 & 0.36 & 0.551 & 0.469 & 0.68 & 0.559 & 0.506 & 0.801 \\
Command R & 0.183 & 0.216 & 0.097 & 0.429 & 0.427 & 0.491 & 0.193 & 0.021 & 0.552 \\
Command R+ & 0.218 & 0.231 & 0.095 & --- & 0.365 & 0.295 & 0.217 & 0.079 & 0.591 \\
Command R7B & 0.153 & 0.14 & 0.085 & 0.166 & 0.158 & 0.529 & 0.172 & 0.027 & 0.462 \\
GPT-4.1 & 0.394 & 0.377 & 0.306 & 0.805 & 0.414 & 0.463 & 0.492 & 0.381 & 0.752 \\
GPT-5 & 0.765 & 0.748 & 0.707 & 0.923 & 0.791 & 0.774 & 0.788 & 0.749 & 0.896 \\
GPT-5 Mini & 0.674 & 0.69 & 0.642 & 0.769 & 0.73 & 0.733 & 0.698 & 0.646 & 0.852 \\
GPT-5 Nano & 0.597 & 0.618 & 0.527 & 0.611 & 0.643 & 0.693 & 0.6 & 0.552 & 0.798 \\
GPT-5.2 & 0.393 & 0.41 & 0.309 & 0.83 & 0.474 & 0.514 & 0.426 & 0.271 & 0.759 \\
Qwen3-235B Instruct & 0.525 & 0.519 & 0.381 & 0.834 & 0.569 & 0.589 & 0.457 & 0.37 & 0.758 \\
Qwen3-235B Thinking & 0.549 & 0.55 & 0.505 & 0.674 & 0.546 & 0.521 & 0.5 & 0.484 & 0.795 \\
Qwen3-80B Instruct & 0.508 & 0.466 & 0.37 & 0.771 & 0.484 & 0.536 & 0.488 & 0.365 & 0.743 \\
Qwen3-80B Thinking & 0.459 & 0.458 & 0.421 & 0.67 & 0.446 & 0.437 & 0.416 & 0.399 & 0.755 \\
DeepSeek-R1 & 0.312 & 0.316 & 0.28 & 0.561 & 0.337 & 0.349 & 0.343 & 0.309 & 0.727 \\
DeepSeek-V3.1 & 0.391 & 0.412 & 0.259 & 0.418 & 0.434 & 0.427 & 0.368 & 0.301 & 0.7 \\
Llama 3.1 405B & 0.232 & 0.316 & 0.292 & 0.527 & 0.362 & 0.344 & 0.318 & 0.235 & 0.647 \\
Llama 3.1 70B & 0.208 & 0.221 & 0.164 & 0.428 & 0.256 & 0.062 & 0.2 & 0.184 & 0.459 \\
Llama 3.1 8B & 0.186 & 0.195 & 0.142 & 0.414 & 0.219 & 0.348 & 0.204 & 0.08 & 0.52 \\
Kimi-K2 Thinking & 0.525 & 0.518 & 0.498 & 0.79 & 0.545 & 0.503 & 0.501 & 0.465 & 0.799 \\
GPT-OSS-120B & 0.593 & 0.597 & 0.541 & --- & 0.663 & 0.7 & 0.616 & 0.537 & 0.824 \\
GPT-OSS-20B & 0.28 & 0.368 & 0.277 & --- & 0.422 & 0.482 & 0.433 & 0.357 & 0.718 \\
GLM-4.5 Air & 0.264 & 0.291 & 0.252 & 0.489 & 0.435 & 0.48 & 0.437 & 0.356 & 0.713 \\
GLM-4.6 & 0.406 & 0.431 & 0.373 & 0.603 & 0.458 & 0.364 & 0.443 & 0.378 & 0.706 \\
\bottomrule
\end{tabular}
\caption{Model scores across scoring schemes for SuperGPQA. Dashes (---) denote models that could not complete the evaluation run in time, and were thus omitted from analyses.}
\label{tab:scores_super_gpqa}
\end{table*}

%% file: appendix/strategy_correlation.tex
\begin{figure*}[t]
    \centering
    \includegraphics[width=\linewidth]{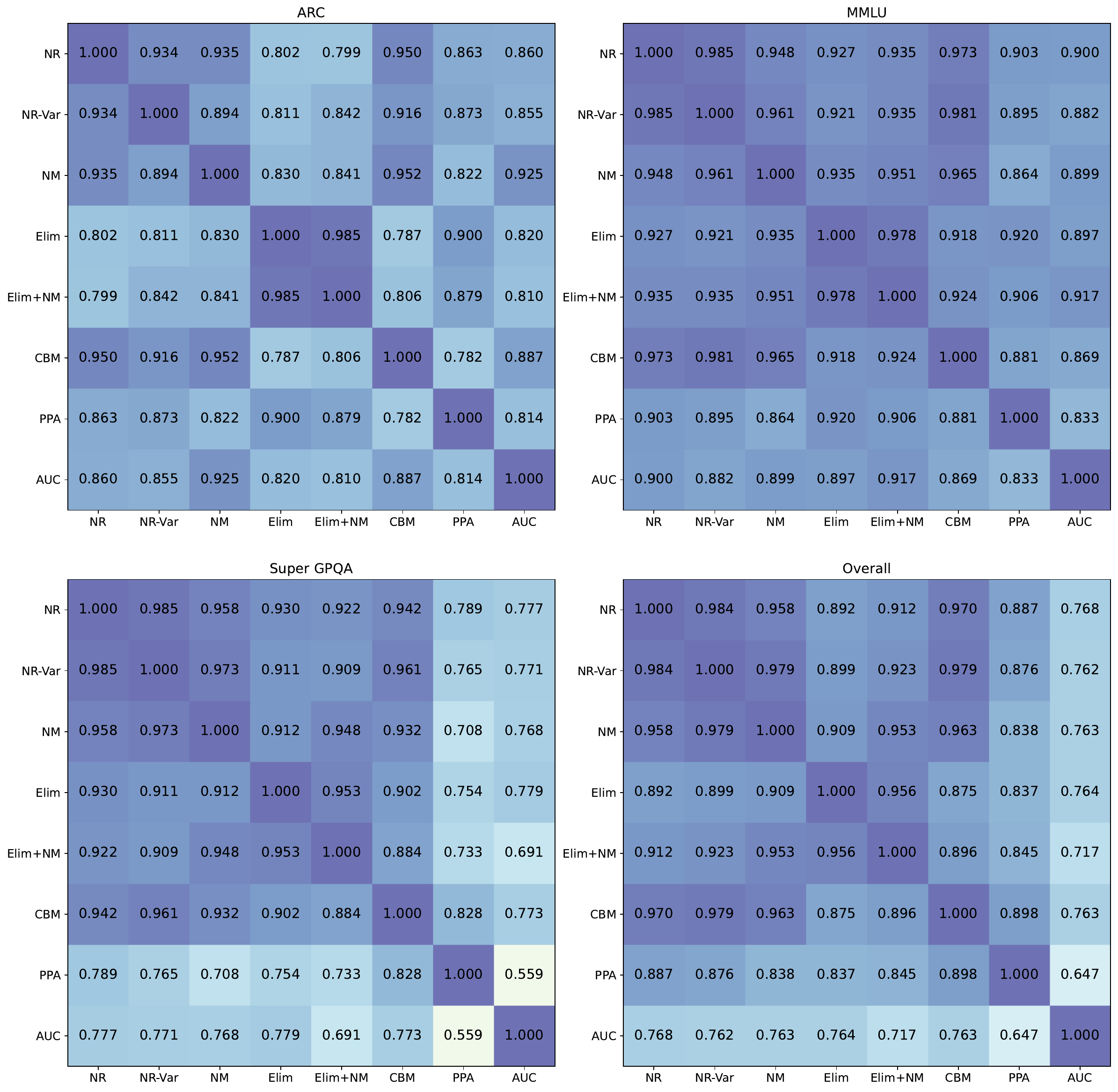}
    \caption{\label{fig:strategy_correlation} Spearman's correlation between model rankings across all scoring schemes and datasets.}
\end{figure*}

\begin{figure*}[t]
    \centering
    \includegraphics[width=\linewidth]{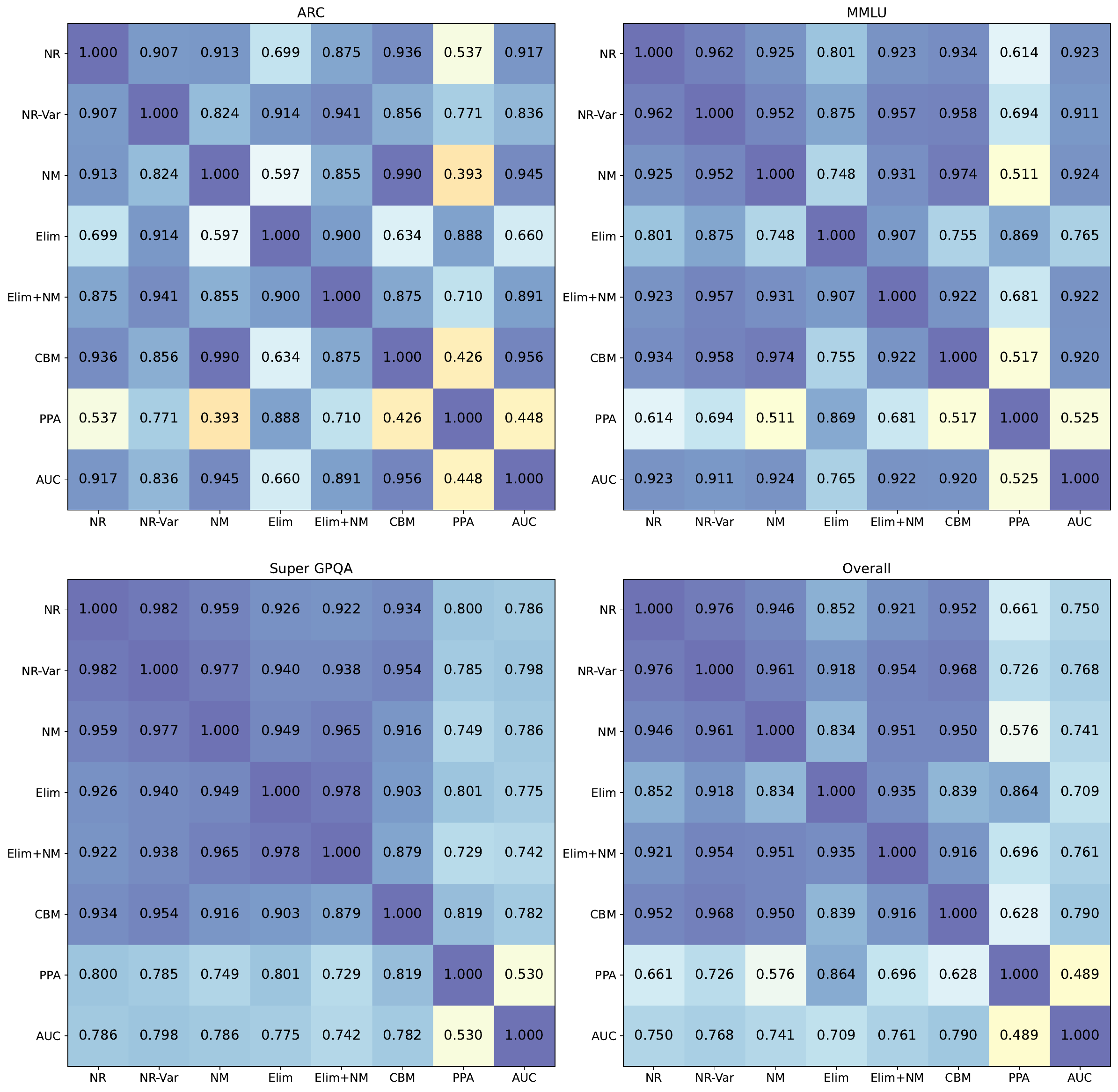}
    \caption{\label{fig:strategy_correlation_scores} Pearson's correlation between scores across all schemes and datasets. We run Min-Max normalization on all scores. When considering disagreements at the score level, the trends in \cref{subsection:ranks} become starker: certain schemes like \Elim, \PPA, and \AUC show large disagreements from \Norm, further suggesting that they test different abilities.}
\end{figure*}

%% file: appendix/llm_arena_all.tex
\begin{figure*}[t]
    \centering
    \includegraphics[width=\linewidth]{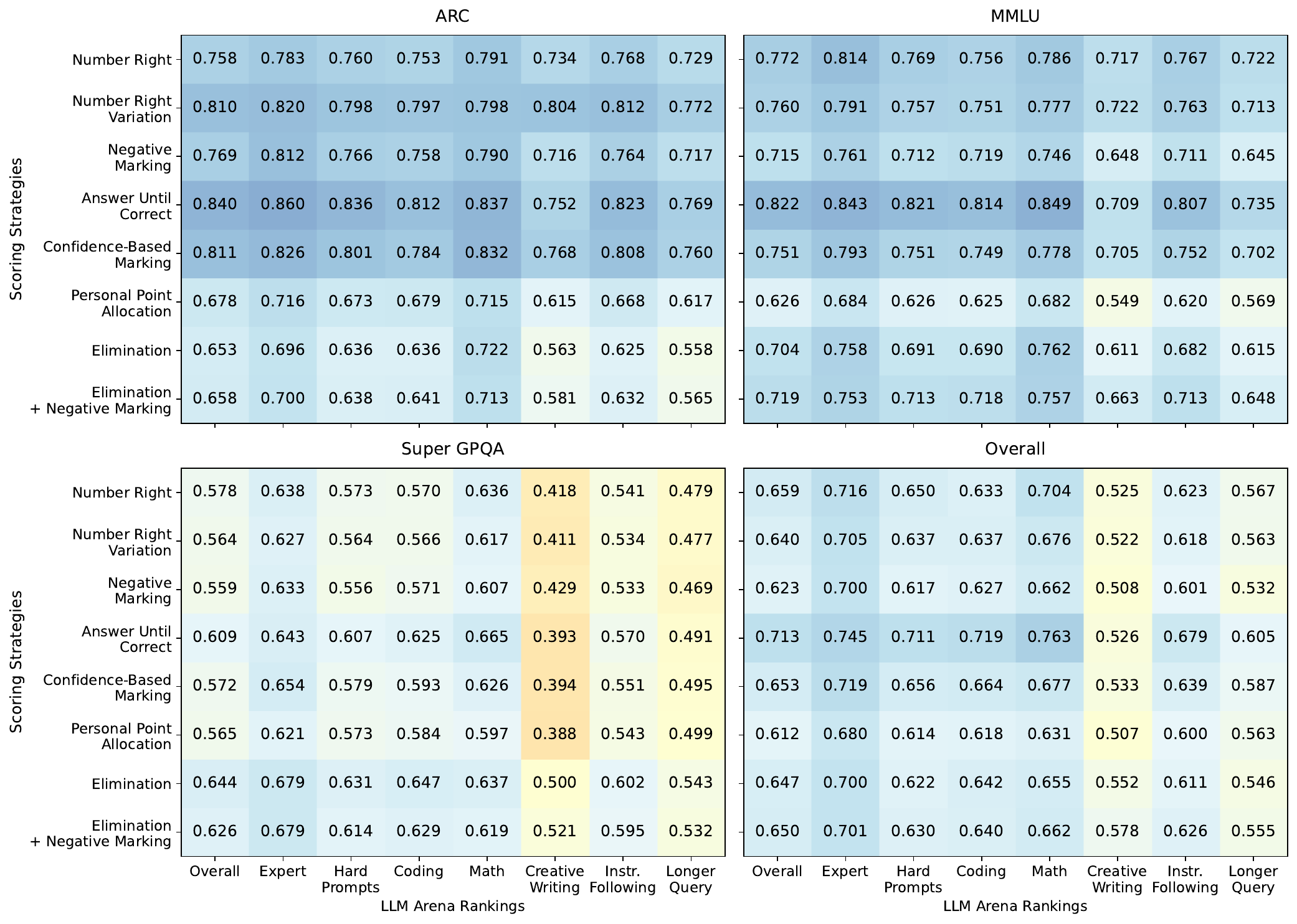}
    \caption{\label{fig:llm_arena_correlation_all} Correlation between scoring strategy and LLM Arena rankings across all datasets. Notably, Answer Until Correct has higher agreement with human preferences on LLM Arena across all datasets.}
\end{figure*}

%% file: appendix/consistency.tex
\begin{figure*}[t]
    \centering
    \includegraphics[width=\linewidth]{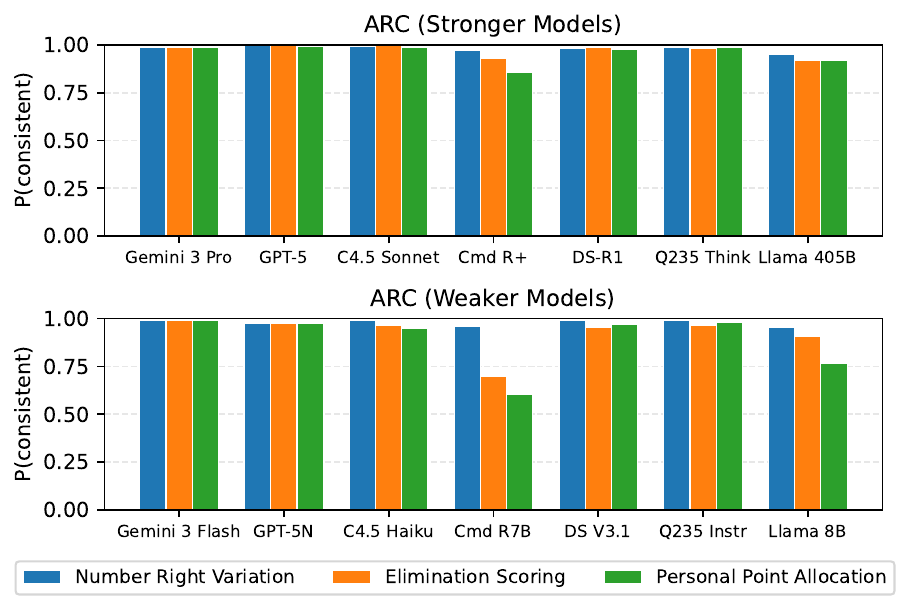}
    \caption{\label{fig:arc_top_consistency} Consistency checks between weaker and stronger models between different scoring regimes and standard scoring on ARC. Weaker models (e.g., smaller, less reasoning) tend to be less consistent, except for strong closed-source models like Gemini and GPT.}
\end{figure*}

\begin{figure*}[t]
    \centering
    \includegraphics[width=\linewidth]{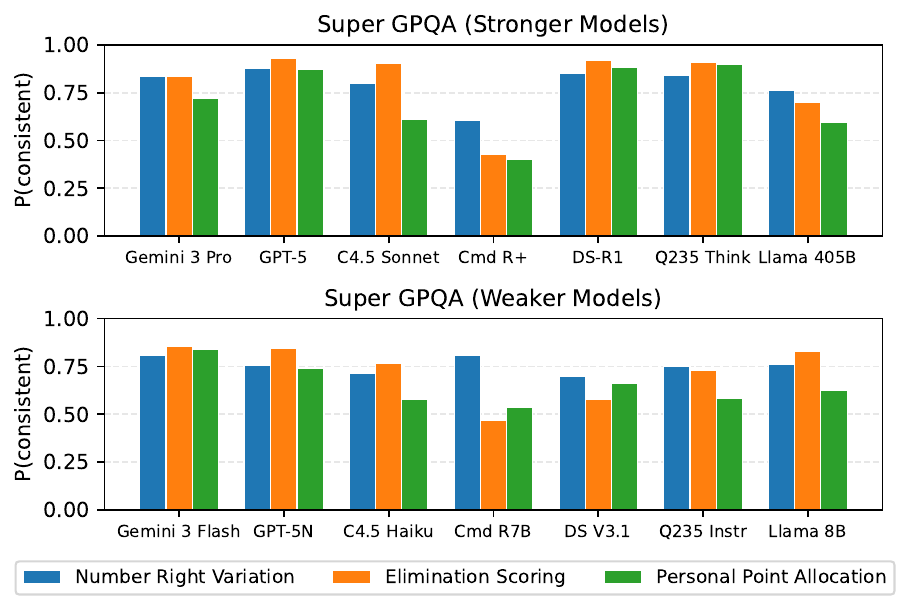}
    \caption{\label{fig:gpqa_top_consistency} Consistency checks between weaker and stronger models between different scoring regimes and standard scoring on SuperGPQA. Weaker models (e.g., smaller, less reasoning) tend to be less consistent, except for strong closed-source models like Gemini and GPT.}
\end{figure*}

\begin{figure*}[t]
    \centering
    \includegraphics[width=\linewidth]{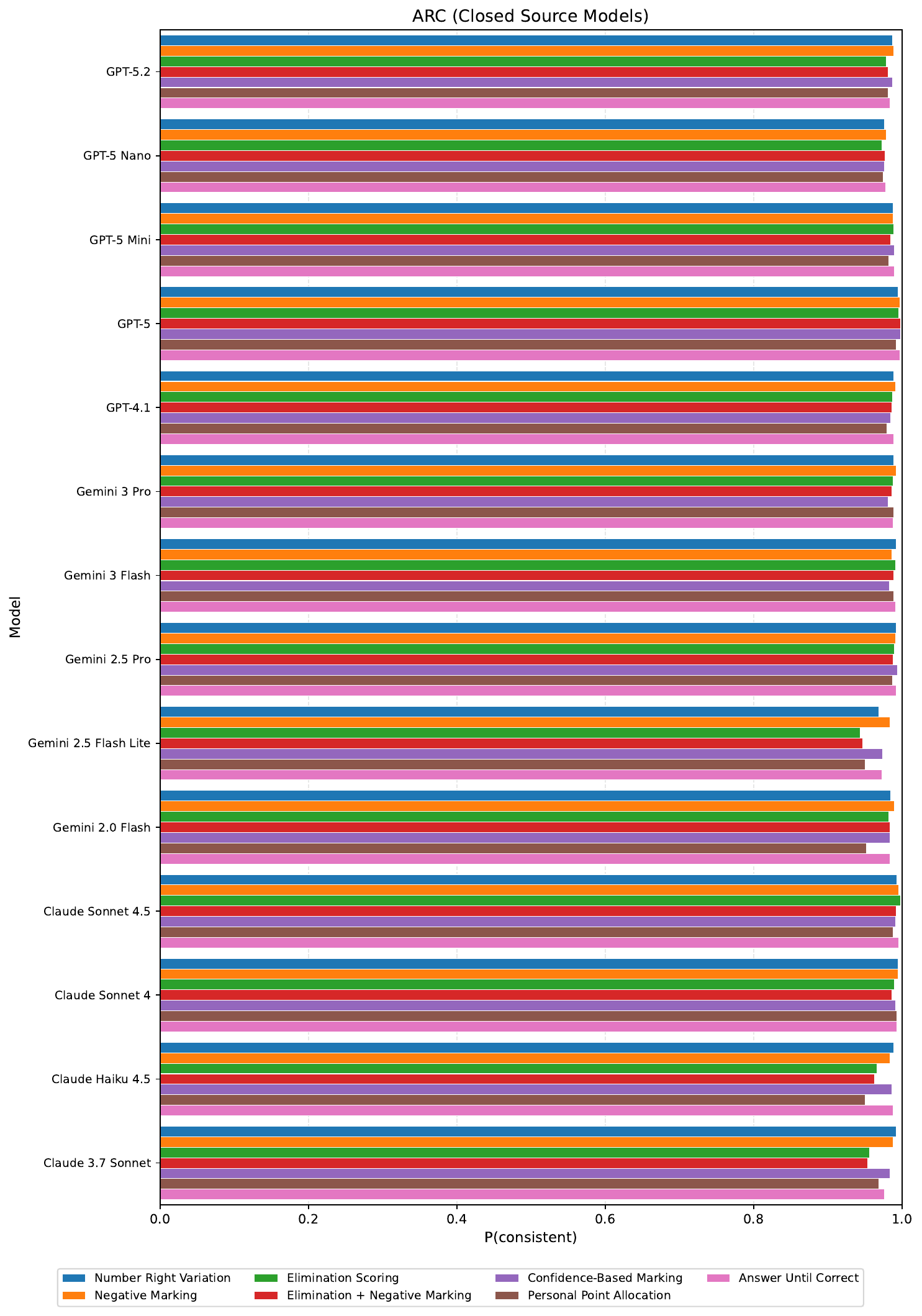}
    \caption{\label{fig:arc_closed_source} Consistency checks for closed-source models on ARC between all scoring strategies.}
\end{figure*}

\begin{figure*}[t]
    \centering
    \includegraphics[width=\linewidth]{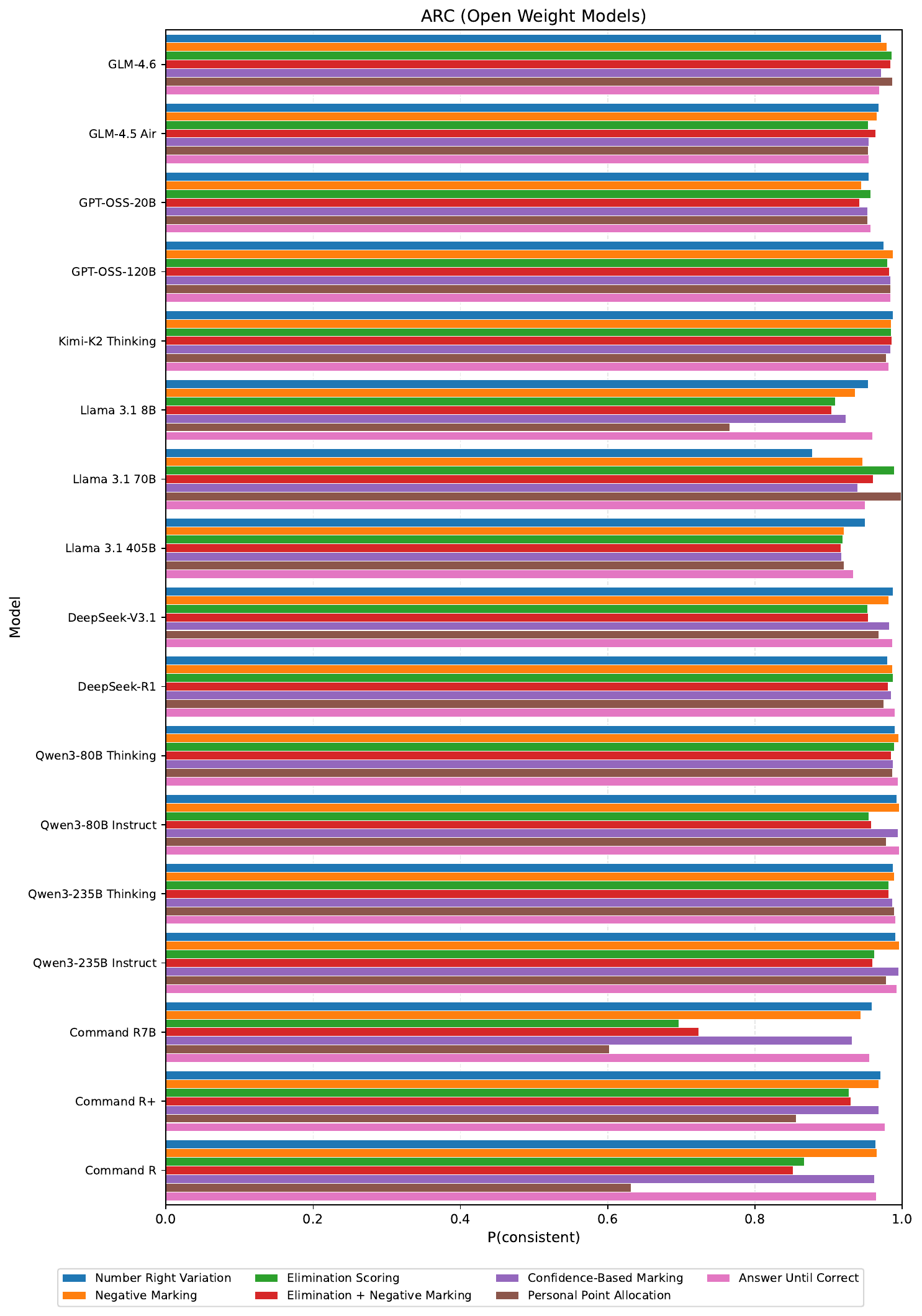}
    \caption{\label{fig:arc_open_weight} Consistency checks for open-weight models on ARC between all scoring strategies.}
\end{figure*}

\begin{figure*}[t]
    \centering
    \includegraphics[width=\linewidth]{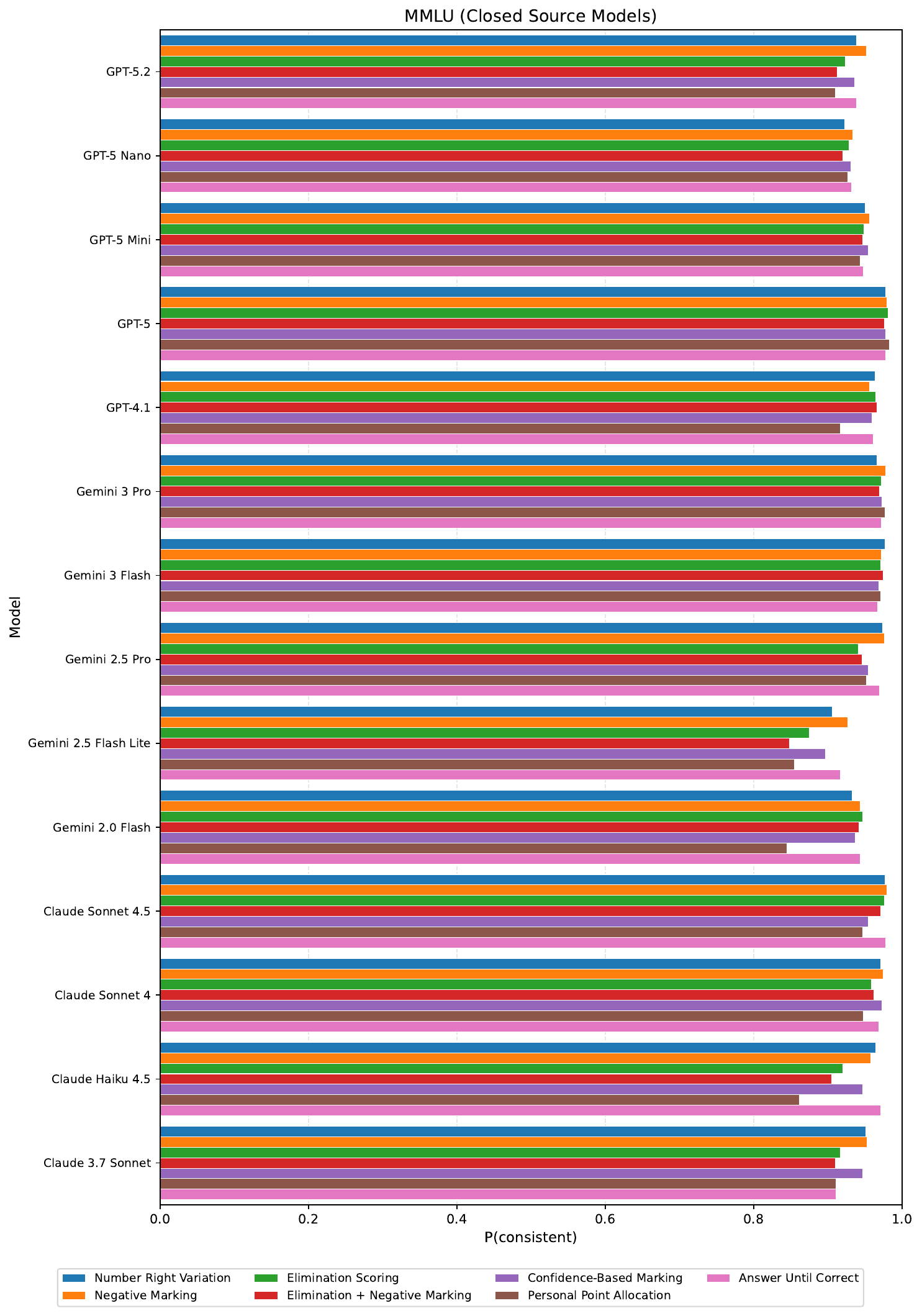}
    \caption{\label{fig:mmlu_closed_source} Consistency checks for closed-source models on MMLU between all scoring strategies.}
\end{figure*}

\begin{figure*}[t]
    \centering
    \includegraphics[width=\linewidth]{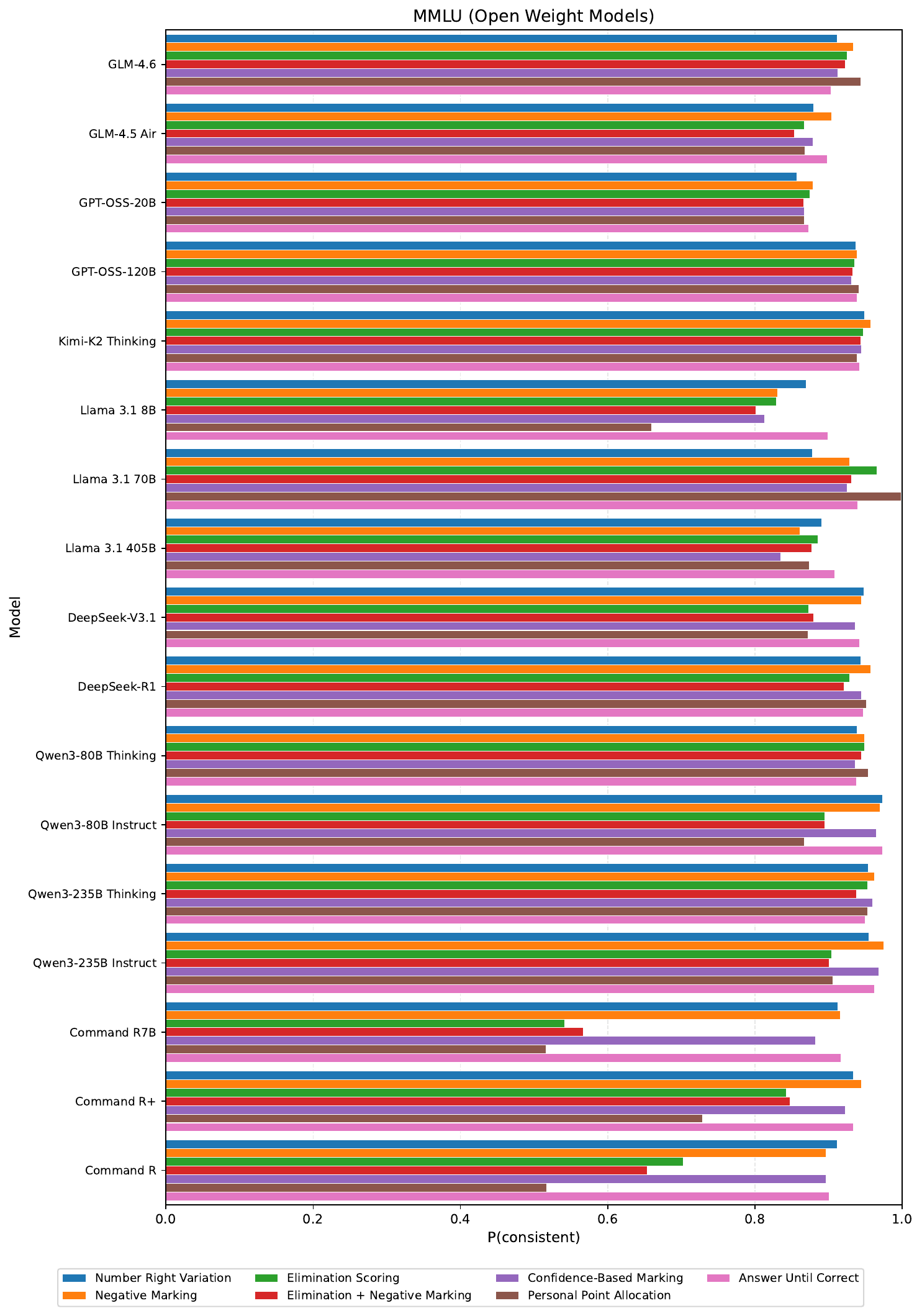}
    \caption{\label{fig:mmlu_open_weight} Consistency checks for open-weight models on MMLU between all scoring strategies.}
\end{figure*}

\begin{figure*}[t]
    \centering
    \includegraphics[width=\linewidth]{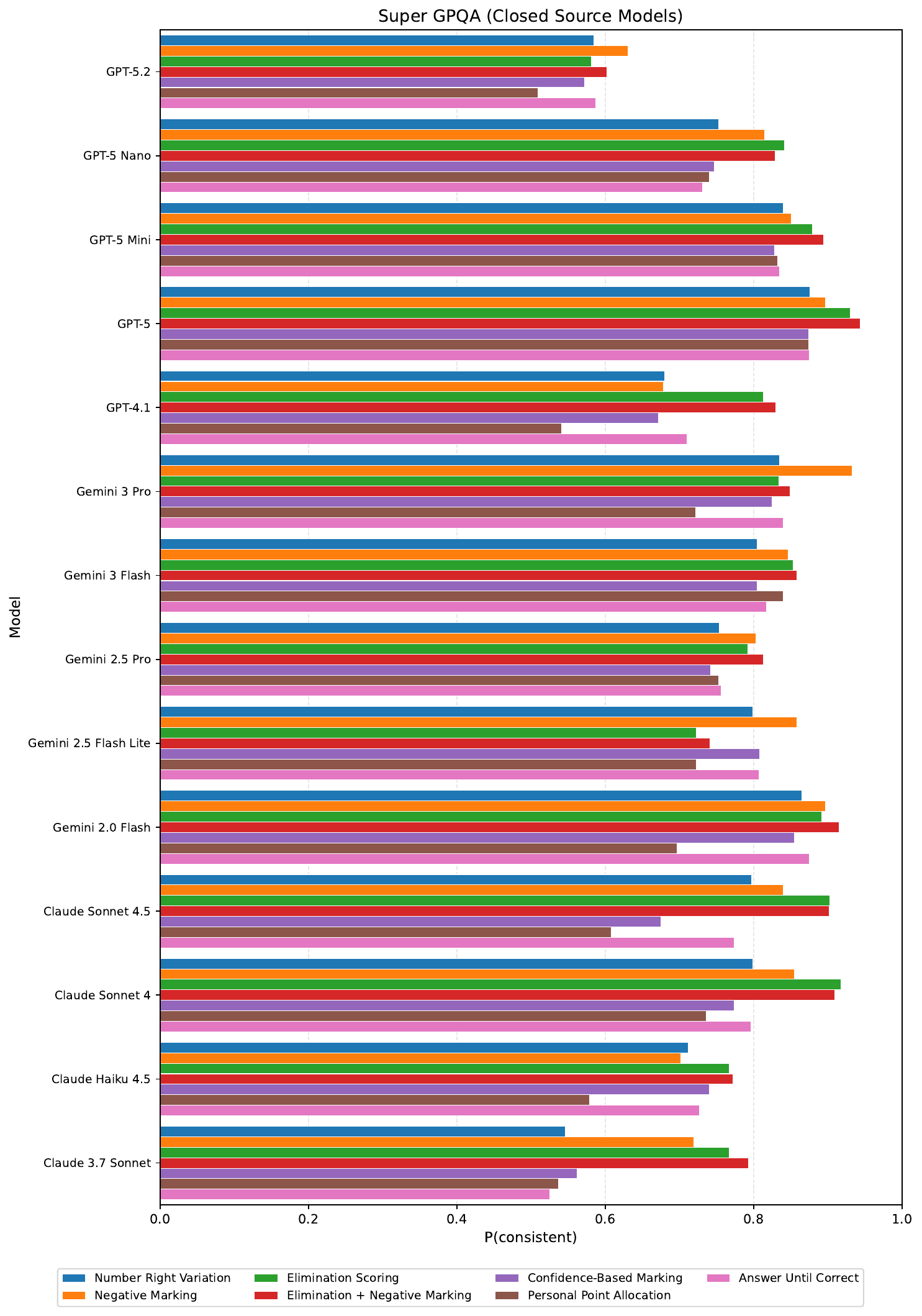}
    \caption{\label{fig:super_gpqa_closed_source} Consistency checks for closed-source models on SuperGPQA between all scoring strategies.}
\end{figure*}

\begin{figure*}[t]
    \centering
    \includegraphics[width=\linewidth]{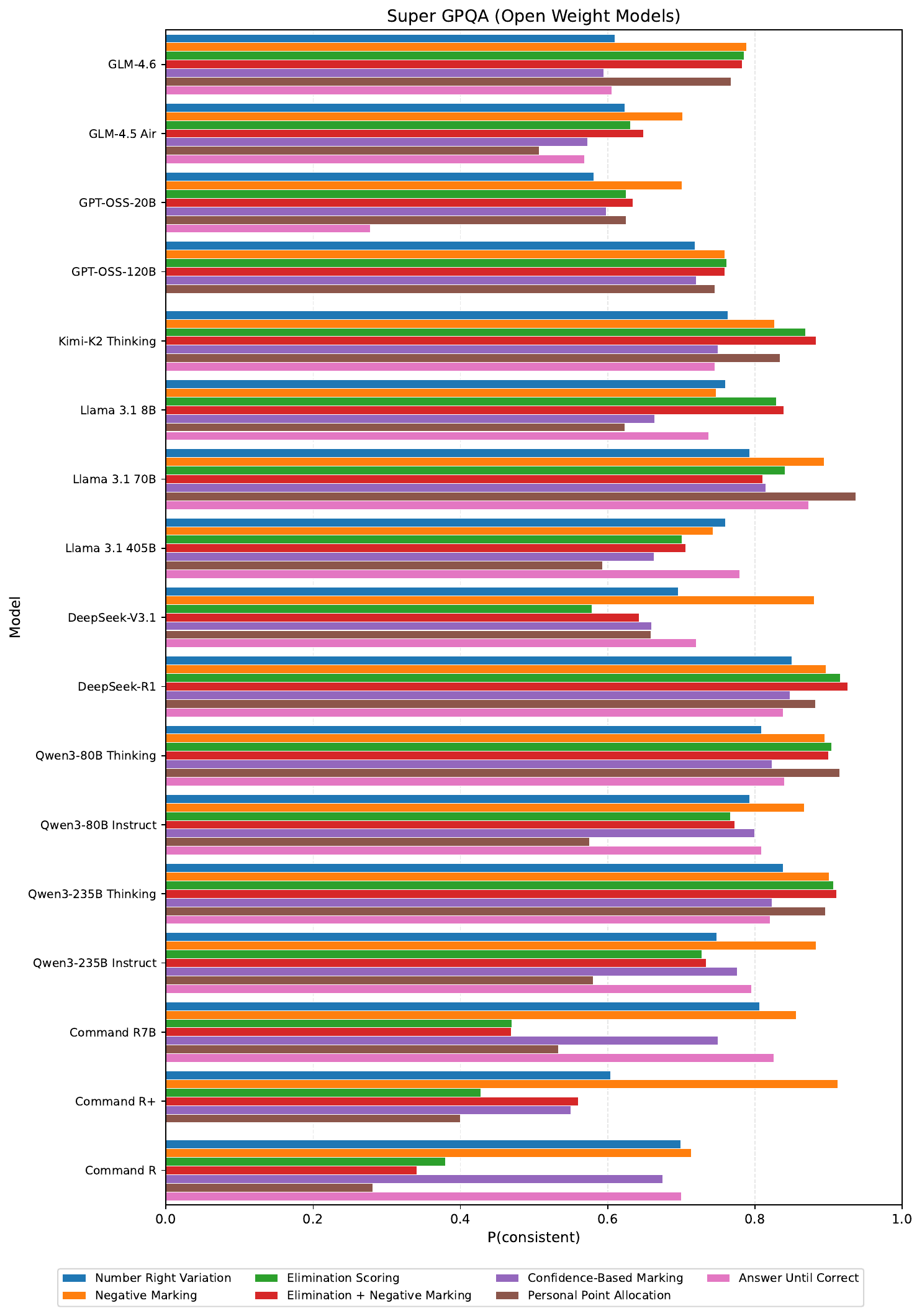}
    \caption{\label{fig:super_gpqa_open_weight} Consistency checks for open-weight models on SuperGPQA between all scoring strategies.}
\end{figure*}

%% file: appendix/behavior.tex
\begin{figure*}[t]
    \centering
    \includegraphics[width=\linewidth]{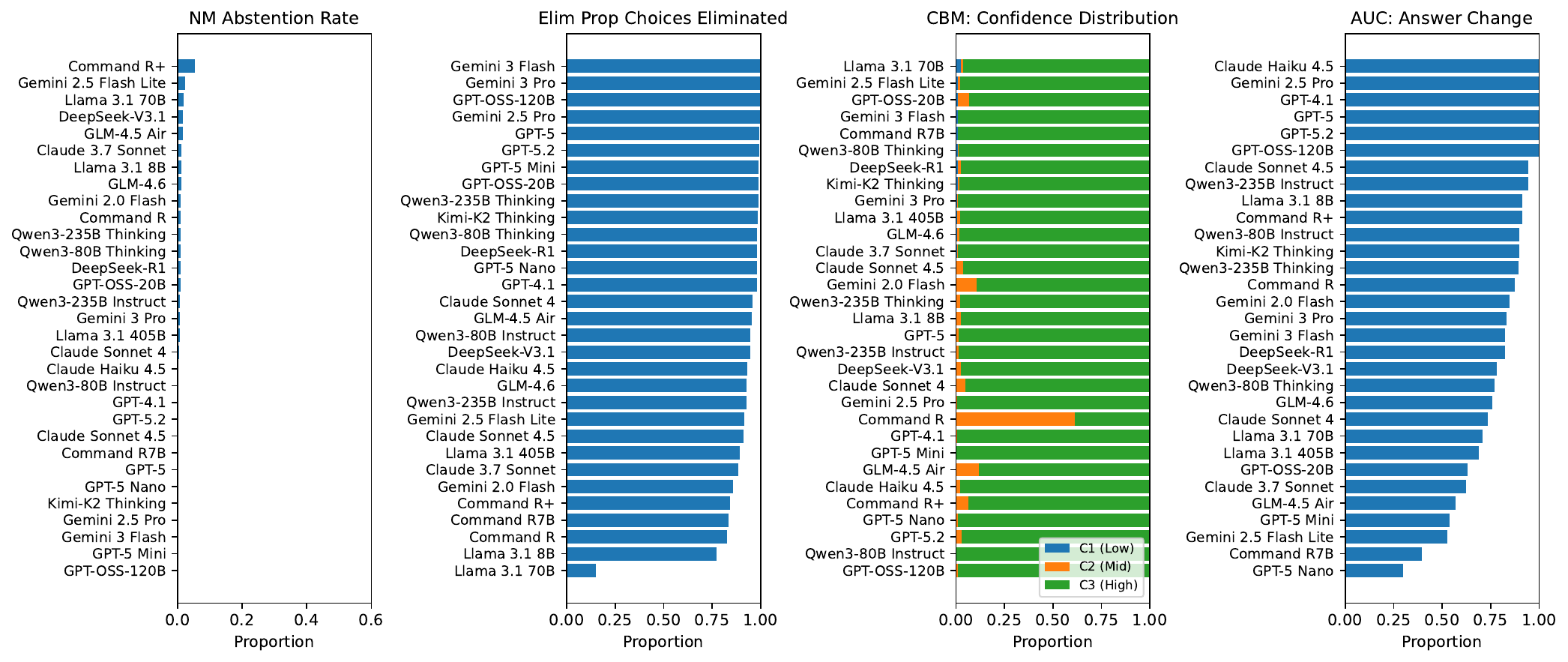}
    \caption{\label{fig:arc_behavior} Model behavior analysis on ARC in Negative Marking, Elimination, Confidence-Based Marking, and Answer Until Correct scoring schemes.}
\end{figure*}

\begin{figure*}[t]
    \centering
    \includegraphics[width=\linewidth]{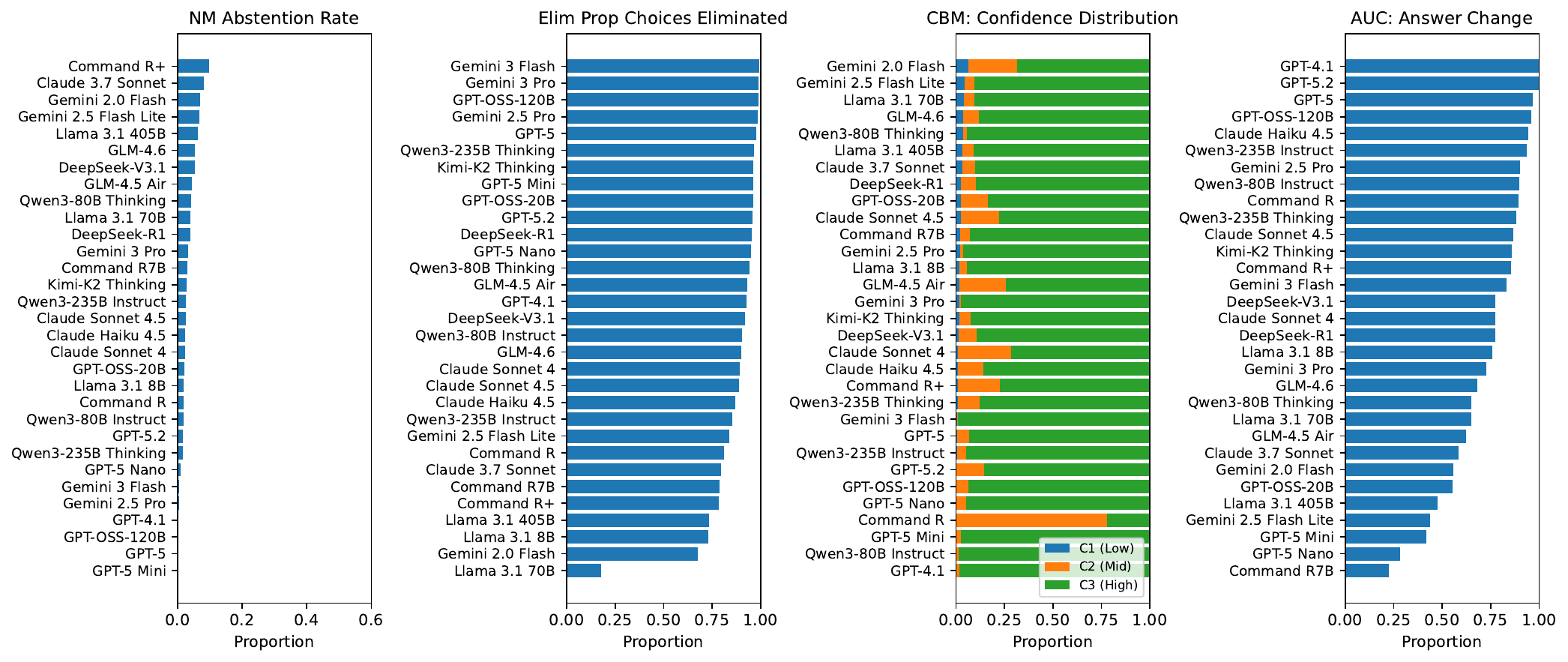}
    \caption{\label{fig:mmlu_behavior} Model behavior analysis on MMLU in Negative Marking, Elimination, Confidence-Based Marking, and Answer Until Correct scoring schemes.}
\end{figure*}

\begin{figure*}[t]
    \centering
    \includegraphics[width=\linewidth]{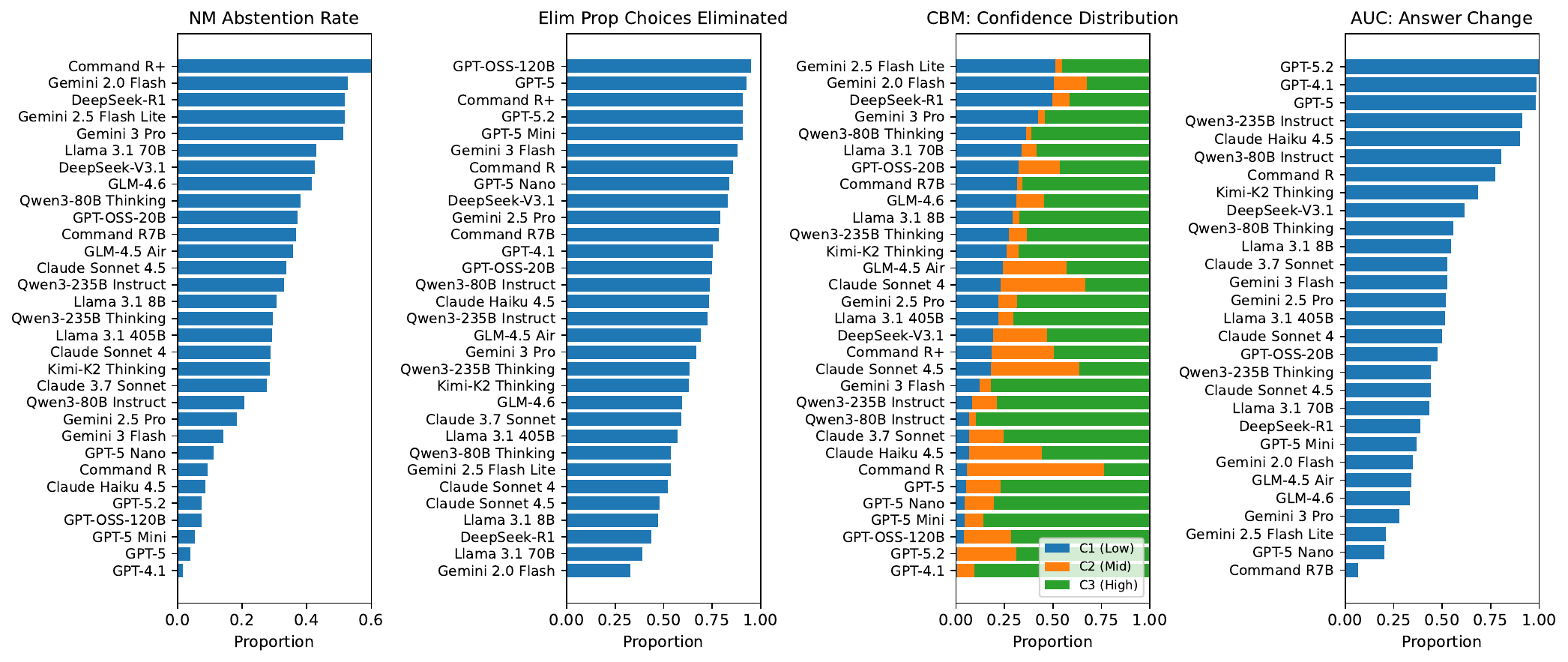}
    \caption{\label{fig:super_gpqa_behavior} Model behavior analysis on Super GPQA in Negative Marking, Elimination, Confidence-Based Marking, and Answer Until Correct scoring schemes.}
\end{figure*}

%% file: appendix/prompts.tex
\hypersetup{
    linkcolor=white,
    citecolor=white,
    urlcolor=white
}

\lstset{
  literate={<}{{<}}1
           {>}{{>}}1
}

\begin{prompt}[title={Prompt \thetcbcounter: Number Right MCQA Prompt}, label=prompt:normal]
<task>\\Answer the following multiple choice question by selecting the letter of the correct answer:

Question: $q$\\
Choices:
$\mathcal{C}$
\\
</task>\\

<scoring>\\
You will receive +1 point if your answer is correct, and 0 points if incorrect.\\
</scoring>\\

<format>\\
Generate your response as a JSON with two keys: 1) "answer" - the letter of the correct answer; and 2) "explanation" - an explanation for your decision. Do not generate anything else.\\
</format>
\end{prompt}

\begin{prompt}[title={Prompt \thetcbcounter: Number Right Variation Prompt}, label=prompt:variation]
<task>\\Please solve the following multiple choice question by choosing the correct option:

Question: $q$\\
Choices:
$\mathcal{C}$
\\
</task>\\

<scoring>\\
Correct answers earn 1 point, while incorrect answers receive 0 points.\\
</scoring>\\

<format>\\
Provide your response as a JSON object containing two fields: 1) "answer" - the letter corresponding to your selected answer; and 2) "explanation" - a brief explanation of your reasoning. Do not generate anything else\\
</format>
\end{prompt}

\begin{prompt}[title={Prompt \thetcbcounter: Negative Marking Prompt}, label=prompt:negative_marking]
<task>\\
Answer the following multiple choice question by selecting the letter of the correct answer, or abstain if you are not confident:\\
Question: $q$\\
Choices:
$\mathcal{C}$
\\
</task>\\

<scoring>\\
You will receive +1 point if your answer is correct, -1/(N-1) points if incorrect (where N is the total number of choices), and 0 points if you abstain.\\
</scoring>\\

<format>\\
Generate your response as a JSON with two keys: 1) "answer" - the letter of the correct answer or "ABSTAIN" if you choose to abstain; and 2) "explanation" - an explanation for your decision. Do not generate anything else\\
</format>

\end{prompt}

\begin{prompt}[title={Prompt \thetcbcounter: Elimination Prompt}, label=prompt:elimination]
<task>\\
Answer the following multiple choice question by eliminating the answers that you think are incorrect:\\
Question: $q$\\
Choices:
$\mathcal{C}$
\\
</task>\\

<scoring>\\
You will receive +1 point for each incorrect answer you eliminate. However, if you eliminate the correct answer, you receive 0 points total (regardless of how many incorrect answers you eliminated). You may eliminate as many or as few answers as you want - there is no requirement to eliminate all answers or any specific number.\\
</scoring>\\

<format>\\
Generate your response as a JSON with two keys: 1) "eliminated" - a list of letters corresponding to the answers you have eliminated (e.g., ["A", "C", "D"]); and 2) "explanation" - an explanation for your eliminations. The "eliminated" list can contain any number of choices (including zero or all choices). Do not generate anything else\\
</format>

\end{prompt}

\begin{prompt}[title={Prompt \thetcbcounter: Elimination with Negative Marking Prompt}, label=prompt:elimination_negative]
<task>\\
Answer the following multiple choice question by eliminating the answers that you think are incorrect:\\
Question: $q$\\
Choices:
$\mathcal{C}$
\\
</task>\\

<scoring>\\
You will receive +1 point for each incorrect answer you eliminate, and -N points for each correct answer you eliminate (where N is the total number of choices). You may eliminate as many or as few answers as you want - there is no requirement to eliminate all answers or any specific number.\\
</scoring>\\

<format>\\
Generate your response as a JSON with two keys: 1) "eliminated" - a list of letters corresponding to the answers you have eliminated (e.g., ["A", "C", "D"]); and 2) "explanation" - an explanation for your eliminations. The "eliminated" list can contain any number of choices (including zero or all choices). Do not generate anything else\\
</format>
\end{prompt}

\begin{prompt}[title={Prompt \thetcbcounter: Confidence Marking Prompt}, label=prompt:confidence]
<task>\\
Answer the following multiple choice question by selecting the letter of the correct answer along with your confidence level:\\
Question: $q$\\
Choices:
$\mathcal{C}$\\
</task>\\

<scoring>\\
You will be scored using Confidence-Based Marking with discrete confidence levels. You must declare your confidence level as C=1 (low), C=2 (mid), or C=3 (high):\\
- C=1 (low confidence): +1 mark if correct, 0 marks (no penalty) if wrong\\
- C=2 (mid confidence): +2 marks if correct, -2 marks if wrong\\
- C=3 (high confidence): +3 marks if correct, -6 marks if wrong\\
Higher confidence levels yield greater rewards for correct answers but also incur greater penalties for incorrect answers. Choose your confidence level carefully based on how certain you are about your answer.\\
</scoring>\\

<format>\\
Generate your response as a JSON with three keys: 1) "answer" - the letter of the correct answer; 2) "confidence" - an integer confidence level of 1 (low), 2 (mid), or 3 (high); and 3) "explanation" - an explanation for your decision. Do not generate anything else\\
</format>
\end{prompt}

\begin{prompt}[title={Prompt \thetcbcounter: Personal Point Prompt}, label=prompt:personal_point]
<task>
Answer the following multiple choice question by allocating confidence scores (weights) to each choice, where all weights should sum to 1.0:

Question: $q$\\
Choices:
$\mathcal{C}$\\
</task>\\

<scoring>\\
You will be scored based on how well your probability distribution matches the true outcome. Your score will be highest when you assign probability 1.0 to the correct answer and 0.0 to all incorrect answers. The more weight you allocate to the correct answer, the higher your score will be. For example, if the correct answer is "B", allocating {{"A": 0.2, "B": 0.6, "C": 0.15, "D": 0.05}} will yield a better score than {{"A": 0.4, "B": 0.3, "C": 0.2, "D": 0.1}} because more weight is placed on the correct answer "B".\\
</scoring>\\

<format>\\
Generate your response as a JSON with two keys: 1) "allocations" - a dictionary mapping each choice letter to its confidence weight (e.g., {{"A": 0.5, "B": 0.3, "C": 0.15, "D": 0.05}}); and 2) "explanation" - an explanation for your allocations. Do not generate anything else\\
</format>
\end{prompt}

\begin{prompt}[title={Prompt \thetcbcounter: Answer Until Correct}, label=prompt:auc]
<task>\\
Answer the following multiple choice question by selecting the letter of the correct answer:\\
Question: $q$\\
Choices:
$\mathcal{C}$\\
</task>\\

<feedback>\\You have already tried answering with the following choices, which were incorrect: $\mathcal{A}'$. Do not select any of these choices as the correct answer.\\</feedback>\\

<scoring>\\
You will have multiple attempts to answer correctly. Your score is calculated as (num\_remaining - 1) / (N - 1), where num\_remaining is the number of choices remaining after your attempts and N is the total number of choices. Answering correctly on your first attempt gives you the maximum score of 1.0. The more attempts you need, the lower your score.\\
</scoring>\\

<format>\\
Generate your response as a JSON with two keys: 1) "answer" - the letter of the correct answer; and 2) "explanation" - an explanation for your decision. Do not generate anything else.\\
</format>

\end{prompt}